\documentclass[10pt,twocolumn,letterpaper]{article}

\usepackage[pagenumbers]{iccv} 

\usepackage{amsmath,amsfonts,bm}

\def\eqref#1{equation~\ref{#1}}

\def\1{\bm{1}}

\def\va{{\bm{a}}}

\def\vx{{\bm{x}}}

\def\vz{{\bm{z}}}

\def\mA{{\bm{A}}}

\def\mH{{\bm{H}}}

\def\mK{{\bm{K}}}

\def\mM{{\bm{M}}}

\def\mO{{\bm{O}}}

\def\mQ{{\bm{Q}}}
\def\mR{{\bm{R}}}

\def\mV{{\bm{V}}}
\def\mW{{\bm{W}}}
\def\mX{{\bm{X}}}

\def\mZ{{\bm{Z}}}

\DeclareMathAlphabet{\mathsfit}{\encodingdefault}{\sfdefault}{m}{sl}
\SetMathAlphabet{\mathsfit}{bold}{\encodingdefault}{\sfdefault}{bx}{n}

\newcommand{\R}{\mathbb{R}}

\definecolor{iccvblue}{rgb}{0.21,0.49,0.74}
\usepackage[pagebackref,breaklinks,colorlinks,allcolors=iccvblue]{hyperref}

\def\paperID{5236} 
\def\confName{ICCV}
\def\confYear{2025}

\title{Beyond Text-Visual Attention: Exploiting Visual Cues for \\ Effective Token Pruning in VLMs}

\author{%
  \textbf{Qizhe Zhang$^{1, 2}$\thanks{Work done during an internship at ByteDance.} \quad Aosong Cheng$^{1}$ \quad Ming Lu$^{3}$ \quad Renrui Zhang$^{4}$ \quad Zhiyong Zhuo$^{1}$} \\
  \textbf{Jiajun Cao$^{1}$ \quad Shaobo Guo$^{2}$ \quad Qi She$^{2}$ \quad Shanghang Zhang$^{1}$\thanks{Corresponding author: \texttt{\{theia, shanghang\}@pku.edu.cn}.}} \\
  $^1$ National Key Laboratory for Multimedia Information Processing,\\
  School of Computer Science, Peking University \\
  $^2$ ByteDance \quad $^3$ Intel Labs China \quad $^4$ CUHK MMLab
}

\begin{document}
\maketitle
\begin{abstract}
Large vision-language models (LVLMs) generally contain significantly more visual tokens than their textual counterparts, resulting in a considerable computational burden. Recent efforts have been made to tackle this issue by pruning visual tokens early within the language model. Most existing works use attention scores between text and visual tokens to assess the importance of visual tokens. However, in this study, we first analyze the text-visual attention in the language model and find that this score is not an ideal indicator for token pruning. Based on the analysis, We propose \textbf{VisPruner}, a plug-and-play method that utilizes visual cues for more effective token pruning in LVLMs. Specifically, we first use visual attention to select a limited number of significant tokens. Then, we remove duplicate tokens from the remaining ones based on their similarity. By retaining diverse tokens alongside the initially selected important tokens, we maximally preserve the visual information of the input image. Experimental results demonstrate that our VisPruner sustains strong performance across various VLM architectures and reduction ratios, significantly outperforming existing methods based on text-visual attention. Notably, without any training, VisPruner can reduce the FLOPs of LLaVA-1.5-7B by 91\% and inference latency by 75\%, while maintaining comparable performance. Our code is available at \href{https://github.com/Theia-4869/VisPruner}{https://github.com/Theia-4869/VisPruner}.
\vspace{-6mm}
\end{abstract}    
\section{Introduction}
\label{sec:introduction}

\definecolor{red}{RGB}{255,0,0}
\definecolor{blue}{RGB}{0,112,192}
\definecolor{green}{RGB}{0,176,80}

\begin{figure}
  \centering
  \includegraphics[width=\linewidth]{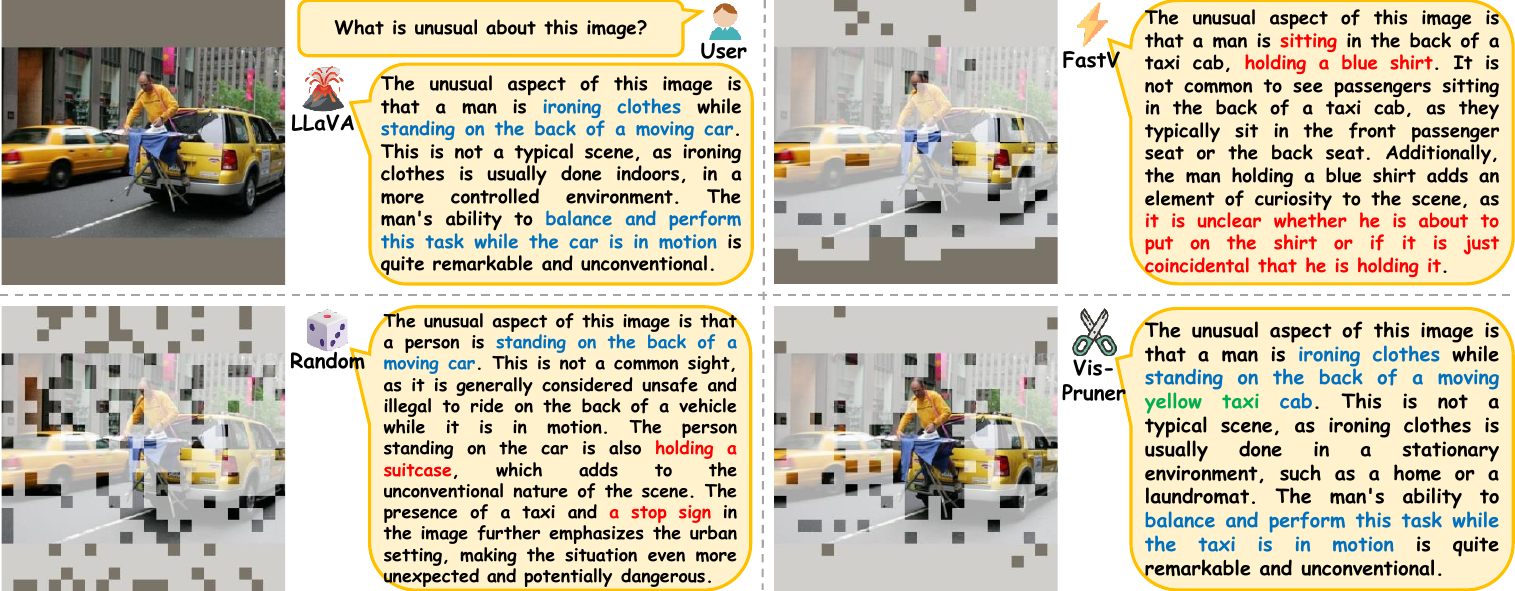}
  \caption{\textbf{The illustration of different pruning methods.} Correct answer parts are shown in \textbf{\textcolor{blue}{blue}}, while hallucinations due to pruning are shown in \textbf{\textcolor{red}{red}}. Text-visual attention methods like FastV often preserve the lower parts of input images, which can lead to the loss of crucial visual information during early pruning (\eg the iron in the man's hand). Random pruning removes positional bias but still fails to preserve important visual information, leading to even more hallucinations (\eg suitcase and stop sign). Our VisPruner, which uses visual cues for pruning, effectively answers the question with a significant token reduction ratio and provides more details (\eg the type and color of the car, shown in \textbf{\textcolor{green}{green}}).}
  \label{fig:case}
  \vspace{-5mm}
\end{figure}

With recent advances in large language models (LLMs)~\cite{ouyang2022instructgpt, touvron2023llama, jiang2023mistral, bai2023qwen}, numerous efforts have been made to extend their powerful reasoning capabilities to multi-modal tasks~\cite{fu2024mme, liu2025mmbench, yu2023mmvet}, giving rise to vision language models (VLMs)~\cite{achiam2023gpt4, team2023gemini, liu2024llava, bai2023qwenvl, chen2024internvl, wang2025cogvlm}. Existing VLMs typically contain a visual encoder (\eg CLIP~\cite{radford2021clip}) to convert visual inputs into sequential representations, a language model (\eg Llama 2~\cite{touvron2023llama2}) to process questions and generate answers, and a modal alignment module (\eg MLP, Q-Former) for the language model to accept visual tokens as input~\cite{liu2024llava, bai2023qwenvl}. Although these approaches have achieved remarkable performance in multi-modal tasks like visual question answering~\cite{goyal2017vqav2, hudson2019gqa, gurari2018vizwiz, lu2022sqa, singh2019textvqa}, the visual tokens, which comprise the majority of the input sequence, significantly increase the computational cost of VLM inference. For instance, LLaVA-1.5~\cite{liu2024llava1.5} accepts $336 \times 336$ input images, which are converted into 576 visual tokens, while their textual counterparts have typically fewer than 100 tokens. Higher input resolutions or video streams lead to even longer visual token sequences (\eg 2880 for LLaVA-NeXT~\cite{liu2024llavanext} and 2560 for  Video-LLaVA~\cite{lin2023videollava}), further escalating computational load.

To reduce the inference cost of VLMs, FastV~\cite{chen2024fastv} initially attempts early pruning of visual tokens within the language model. It identifies that, in the deeper layers of the language model, a substantial number of visual tokens receive significantly less attention than their textual counterparts, and discards many of these low-attention visual tokens after layer 2. Subsequently, abundant recent work follows this paradigm, pruning visual tokens based on text-visual attention from the language model~\cite{ye2024fitprune, zhang2024sparsevlm, xing2024pyramiddrop}. Although visual tokens in VLM are redundant, we find that \textbf{\textit{text-visual attention within the language model may not be an ideal indicator for token pruning}}. As illustrated in ~\cref{fig:case}, pruning based on text-visual attention suffers from positional bias, primarily preserving the lower parts of the image with less useful information (or even completely ineffective paddings). Meanwhile, tokens in the center region with crucial visual information (\eg the iron) are discarded, making it difficult for the model to answer correctly.

We conduct an in-depth analysis of text-visual attention in the language model and make two findings: (1) \textbf{Text-visual attention shift.} Affected by the long-term decay property of rotary position embedding in the language model~\cite{su2024roformer}, for inputs structured as [\textit{image, text}], text tokens located later tend to pay higher attention towards visual tokens that are physically closer to them (\ie patches at the lower part of the input image), a phenomenon that exists from the first layer~\cite{hong2024rope}. (2) \textbf{Text-visual attention dispersion.} Although adjusting the position embedding of the visual token part can avoid the shift issue, both before and after the modification, the same problem persists: the attention density curve exhibits high entropy and low peak, resembling a more uniform distribution, making it difficult to select actually important visual tokens. While these phenomena do not impact the inference process with full visual tokens, pruning based on text-visual attention significantly degrades performance, especially at higher reduction ratios. 

Based on our observations, we propose \textbf{VisPruner}, a plug-and-play approach that exploits visual cues for more effective token pruning in VLMs. Our VisPruner first selects a small number of important tokens with rich visual information based on the attention of the visual encoder. Then, to ensure that the retained tokens do not focus solely on the foreground and overlook background information, we eliminate duplicate tokens from the remainder based on similarity. The remaining diverse tokens, along with the initially selected important tokens, maximize the retention of visual information in the input image. Notably, unlike text-visual attention-based methods that prune within the language model, our approach leverages visual cues to prune before the language model, ensuring compatibility with various attention optimization techniques (\eg FlashAttention~\cite{dao2022flashattention}) without the need for full attention within the language model, thus achieving higher inference efficiency.

Benefiting from its simple design, we can easily apply VisPruner to different VLMs, including the LLaVA series~\cite{liu2024llava1.5, liu2024llavanext, lin2023videollava} and other architectures like Qwen-VL~\cite{bai2023qwenvl}. Extensive experiments on various vision-language benchmarks demonstrate that VisPruner consistently outperforms other text-visual attention-based methods under different reduction ratios. For example, VisPruner is capable of pruning \textbf{94.4\%} of visual tokens in LLaVA-1.5-7B, reducing inference FLOPs by more than \textbf{95\%}, and maintaining \textbf{91.5\%} of the original performance across 10 tasks without any additional training. In scenarios involving longer visual token sequences, such as high-resolution images (LLaVA-NeXT) or video inputs (Video-LLaVA), VisPruner continues to achieve superior performance with its simple design. In summary, the contributions of our work are three-fold:

\begin{itemize}[10pt]
    \item[1.] We conduct a thorough investigation of text-visual attention in the language model and discover that this attention is not an ideal indicator for token pruning.

    \item[2.] We introduce \textbf{VisPruner}, a plug-and-play approach that exploits visual cues for more effective token pruning in VLMs based on our observations.

    \item[3.] We apply VisPruner to various VLMs and conduct extensive experiments across diverse vision-language tasks, demonstrating our method can maintain high performance even under extreme reduction ratios.
\end{itemize}

\section{Related Work}
\label{sec:related_work}

\noindent \textbf{Vision-language models (VLMs).} The recent impressive success of large language models (LLMs)~\cite{ouyang2022instructgpt, touvron2023llama, jiang2023mistral, bai2023qwen} leads to a trend of extending their powerful reasoning capabilities to multi-modal comprehension tasks, giving birth to vision-language models (VLMs)~\cite{achiam2023gpt4, team2023gemini}. These VLMs generally include a visual encoder for serializing representations of input images and a language model for text generation. To make the language model acceptable for visual representations as input, they typically align the visual and language modalities with an alignment module, which can be a simple linear layer~\cite{liu2024llava}, an MLP projector~\cite{liu2024llava1.5}, or a deep query-based network~\cite{li2023blip2, bai2023qwenvl}. Although this allows the language model to have visual perception, the introduction of long visual token sequences increases the computational burdens. Furthermore, studies have shown that existing VLMs still suffer from visual shortcomings~\cite{tong2024shortcoming} or hallucinations~\cite{huang2024hallucination}. To mitigate this, efforts have been made to improve their performance by increasing the resolution of input images~\cite{luo2024llavahr, xu2024llavauhd}, which further exacerbates the computational burden. Optimizing the inference process of VLMs is an urgent task to enable their application in resource-constrained real-world scenarios.

\noindent \textbf{Token reduction for VLMs.} One way to optimize VLM inference is by reducing the visual tokens that occupy the majority of the input sequence. Several studies have explored token reduction in language models~\cite{dai2020funneltransformer, huang2022pyramidbetr, nawrot2022dynamicpooling, rae2019compressivetransformer}. Compared with text, image information tends to have higher redundancy, making visual token reduction for VLMs more reasonable and effective. Various efforts have been made to compress visual tokens into a more compact representation~\cite{shang2024llavaprumerge, chen2024llavolta, li2024llamavid, zhang2025llavamini, li2024tokenpacker, hu2024mqt, cai2024m3}. However, these methods require additional training, increasing computational burden and training costs. In this work, we mainly focus on training-free VLM inference optimization, typically achieved by removing the less important tokens. FastV~\cite{chen2024fastv} first identifies the redundancy and inefficiency of visual tokens in VLMs and proposes a simple method to remove visual tokens with low attention scores after the second layer of the language model. SparseVLM~\cite{zhang2024sparsevlm} removes distractions from text prompts and uses more accurate text attention to progressively sparsify visual tokens. After this, an increasing number of studies choose to rely on text-visual attention within the language model to assess the importance of visual tokens~\cite{ye2024fitprune, xing2024pyramiddrop}. However, after an in-depth investigation, we find that such text-visual attention may not be an ideal indicator for token pruning. Unlike these previous work, we propose to exploit visual cues for more effective token pruning. Notably, some concurrent work shares similar conclusions with us~\cite{zhao2024sgl, wang2024vtccls, yang2024visionzip}.
\section{Text-Visual Attention Investigation}
\label{sec:analysis}

Most prior research utilizes text-visual attention in the language model to evaluate the significance of visual tokens for pruning. In this section, we conduct an in-depth investigation and suggest that this attention may not be an ideal indicator for token pruning. We begin in \cref{sec3:p1} by introducing the attention mechanisms in VLMs as preliminaries. Then, in \cref{sec3:p2} and \cref{sec3:p3}, we analyze two phenomena observed in text-visual attention, termed \textbf{attention shift} and \textbf{attention dispersion}, which are influenced by positional bias and density distribution, respectively.

\subsection{Preliminaries}
\label{sec3:p1}

Existing VLMs generally consist of two core components: a visual encoder and a language model, both are the Transformer-based architecture~\cite{vaswani2017transformer, dosovitskiy2020vit}. Although they both rely on the self-attention mechanism, there are subtle differences in specific implementations. The visual encoder (\eg CLIP~\cite{radford2021clip}) employs a global attention mechanism, where the patch-wise image $\mX = [{\vx}_{\text{cls}}; {\vx}_{\text{img}}^{1}, {\vx}_{\text{img}}^{2}, \cdots, {\vx}_{\text{img}}^{n}] \in {\R}^{(n+1) \times d}$ is first converted into the query $\mQ$, key $\mK$ and value $\mV$ through three weight matrices ${\mW}_Q, {\mW}_K, {\mW}_V \in {\R}^{d \times d}$ respectively:
\begin{equation}
  \mQ = \mX {\mW}_Q, \quad \mK = \mX {\mW}_K, \quad \mV = \mX {\mW}_V
  \label{eq:qkv}
\end{equation}
where $n$ is the length of the image token sequence, and $d$ is the size of the hidden state. Then, the scaled dot-product attention is calculated as follows:
\begin{equation}
  \mA = \text{softmax}\left(\frac{\mQ {\mK}^{T}}{\sqrt{d}}\right), \quad \mO = \mA \mV
  \label{eq:self-attention}
\end{equation}
Here, we refer to the first row of $\mA$ as \textit{[\texttt{CLS}] attention}.

While the language model (\eg Llama 2~\cite{touvron2023llama2}) utilizes causal self-attention~\cite{vaswani2017transformer}, where each token can only attend to the previous tokens (\ie those from the past) and not the following ones (\ie those from the future). For the whole input sequence $\mX = [{\vx}_{\text{sys}}; {\vx}_{\text{img}}; {\vx}_{\text{txt}}] \in {\R}^{l \times d}$, the query $\mQ$, key $\mK$ and value $\mV$ are obtained in the same way, and the causal attention with relative rotary position embedding is then calculated as:
\begin{equation}
  \mA = \text{softmax}\left(\frac{({\mR}_{\theta}\mQ) {({\mR}_{\theta}{\mK})}^{T} + \mM}{\sqrt{d}}\right), \ \mO = \mA \mV
\label{eq:causal-attention}
\end{equation}
where $l$ is the total length of the input sequence (including system prompt, image patches and text question), ${\mR}_{\theta} \in {\R}^{d \times d}$ is the block-diagonal rotation matrix, and $\mM \in {\R}^{l \times l}$ is a lower triangular causal mask ensuring that each token attends only to itself and previous tokens. We refer to the attention received by visual tokens received from the last text token as \textit{last attention} following \cite{chen2024fastv}. And multi-head attention is averaged across different heads for analysis.

\subsection{Text-visual attention shift}
\label{sec3:p2}

\begin{figure*}
  \centering
  \begin{subfigure}{0.325\linewidth}
    \includegraphics[width=\linewidth]{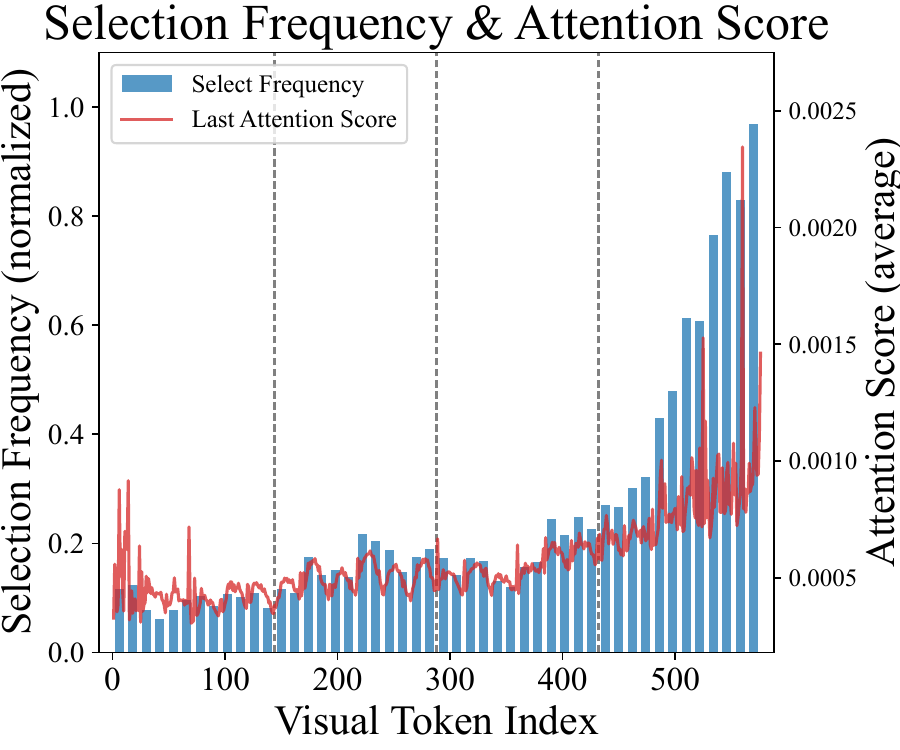}
    \caption{}
    \label{fig:frequency_attention}
  \end{subfigure}
  \hfill
  \begin{subfigure}{0.325\linewidth}
    \includegraphics[width=\linewidth]{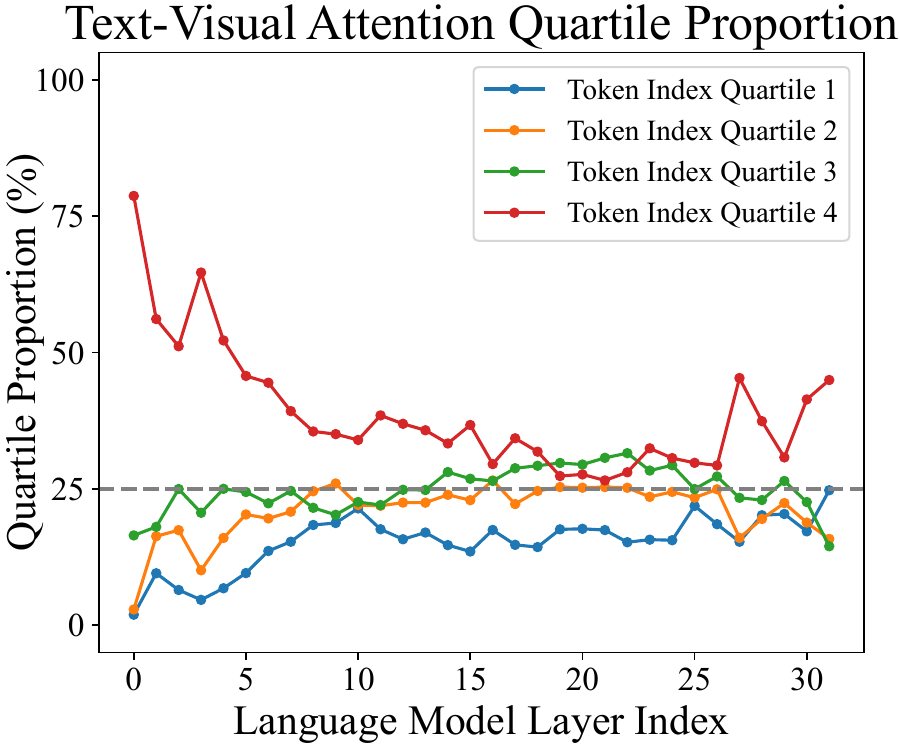}
    \caption{}
    \label{fig:last_attn_quartile}
  \end{subfigure}
  \hfill
  \begin{subfigure}{0.325\linewidth}
    \includegraphics[width=\linewidth]{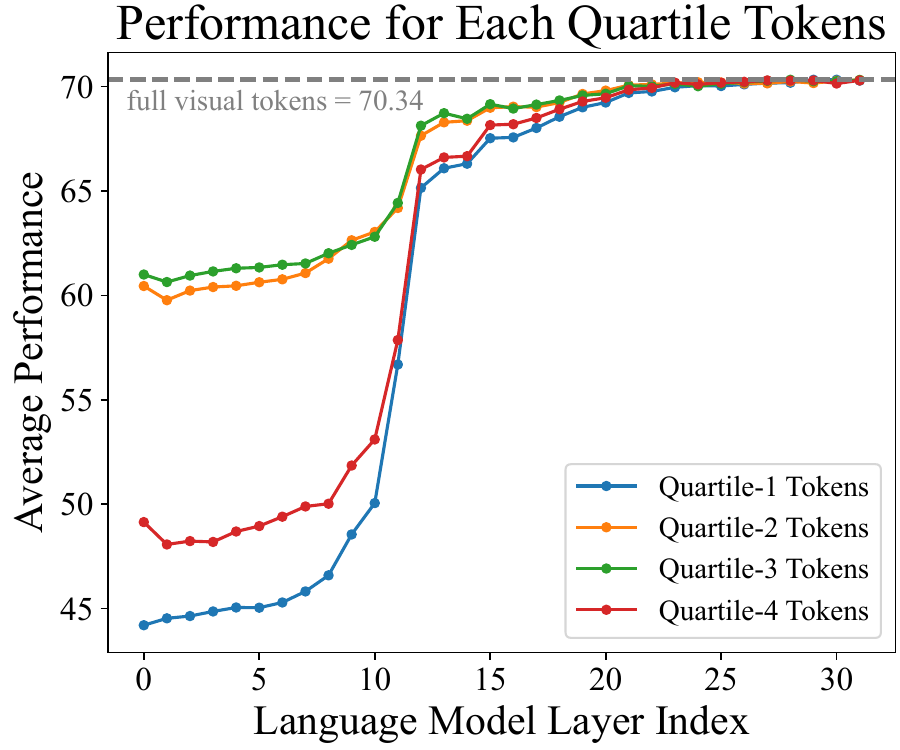}
    \caption{}
    \label{fig:performance_quartile}
  \end{subfigure}
  \caption{\textbf{Analysis of text-visual attention shift.} (a) \textbf{Selection frequency and received attention of visual tokens.} There is a clear positive correlation between selection frequency and the attention received, which is also accompanied by a positional bias. (b) \textbf{Proportion of visual tokens with top 25\% text-visual attention across each quartile position.} In the shallower layers of the language model, attention is notably concentrated on visual tokens with larger indices. (c) \textbf{Performance with only visual tokens from each quartile position.} In the shallower layers, retaining visual tokens located in the central region yields higher performance.}
  \label{fig:attn_shift}
  \vspace{-2mm}
\end{figure*}

To analyze text-visual attention in VLMs, we first randomly sample $N$ image-text pairs from the LLaVA-mix665k data~\cite{liu2024llava1.5}. These samples are then used to prompt the VLM to generate responses, during which we extract the attention received by the visual tokens. In this section, $N$ is set to 1,000, and we use LLaVA-1.5-7B as the VLM.

Following FastV~\cite{chen2024fastv}, we prune 75\% visual tokens after layer 2 of the language model based on text-visual attention and visualize the frequency with which visual tokens at various positions are selected to be retained, as well as their average attention received across $N$ samples. As shown in \cref{fig:frequency_attention}, there is a clear positive correlation between the selection frequency and the attention received, along with a noticeable shift in text-visual attention. Text tokens tend to focus on visual tokens with higher indices, due to the long-term decay property of rotary position embedding~\cite{su2024roformer}. In the input sequence of the language model, text tokens are located after visual tokens, thus text tokens exhibit higher attention towards later-positioned visual tokens (\ie closer to themselves). We demonstrate the relationship between this shift and the language model layer index in \cref{fig:last_attn_quartile}. It can be seen that positional bias appears from the first layer and is more pronounced in the shallower layers, aligning with conclusions from previous work~\cite{hong2024rope}.

A natural question then arises: \textbf{In the presence of inherent positional bias, do higher attention scores necessarily represent richer visual information?} To answer this question, we separately preserve visual tokens from each quartile position and evaluate the average performance on four benchmarks of GQA~\cite{hudson2019gqa}, TextVQA~\cite{singh2019textvqa}, POPE~\cite{li2023pope}, and MME~\cite{fu2024mme} at different layers, as shown in \cref{fig:performance_quartile}. To our surprise, in the shallower layers of the model, visual tokens located in the central region of the image maintain the highest performance, rather than those positioned later, which received the most text attention. The difference between these two diminishes as the positional bias weakens beyond the tenth model layer. However, optimizing inference efficiency requires us to prune visual tokens as early as possible within the language model, indicating that the phenomenon of text-visual attention shift indeed has a negative impact on performance after pruning.

\subsection{Text-visual attention dispersion}
\label{sec3:p3}

\begin{figure}
  \centering
  \begin{subfigure}{0.24\linewidth}
    \includegraphics[width=\linewidth]{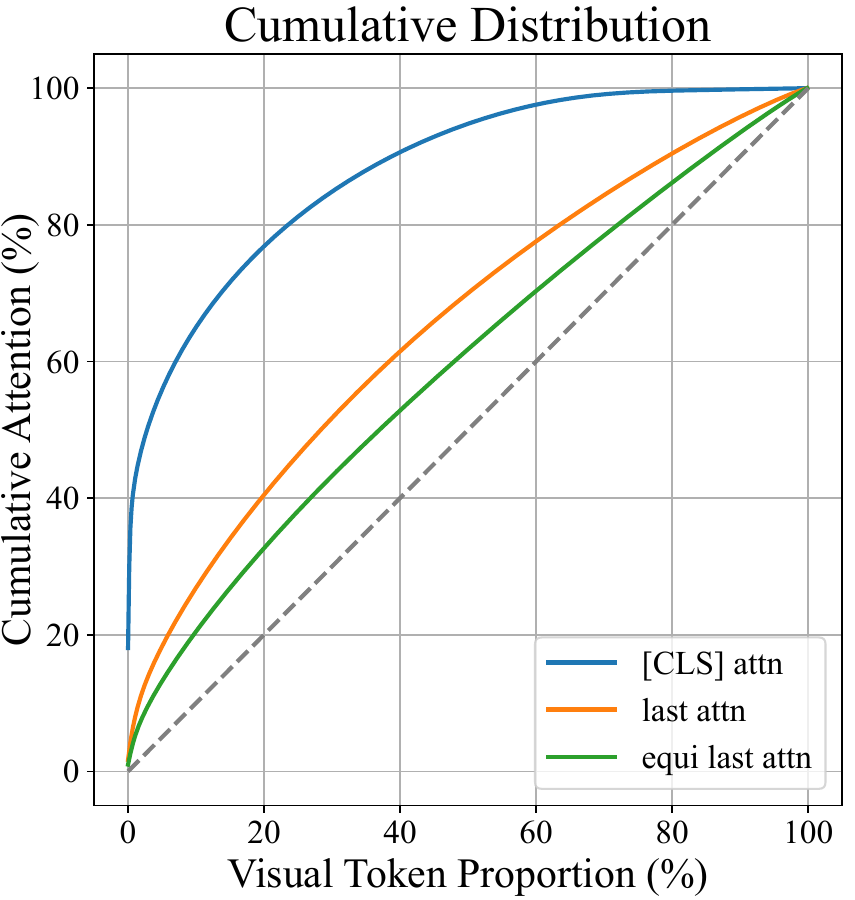}
    \caption{}
    \label{fig:CDF}
  \end{subfigure}
  \hfill
  \begin{subfigure}{0.75\linewidth}
    \includegraphics[width=\linewidth]{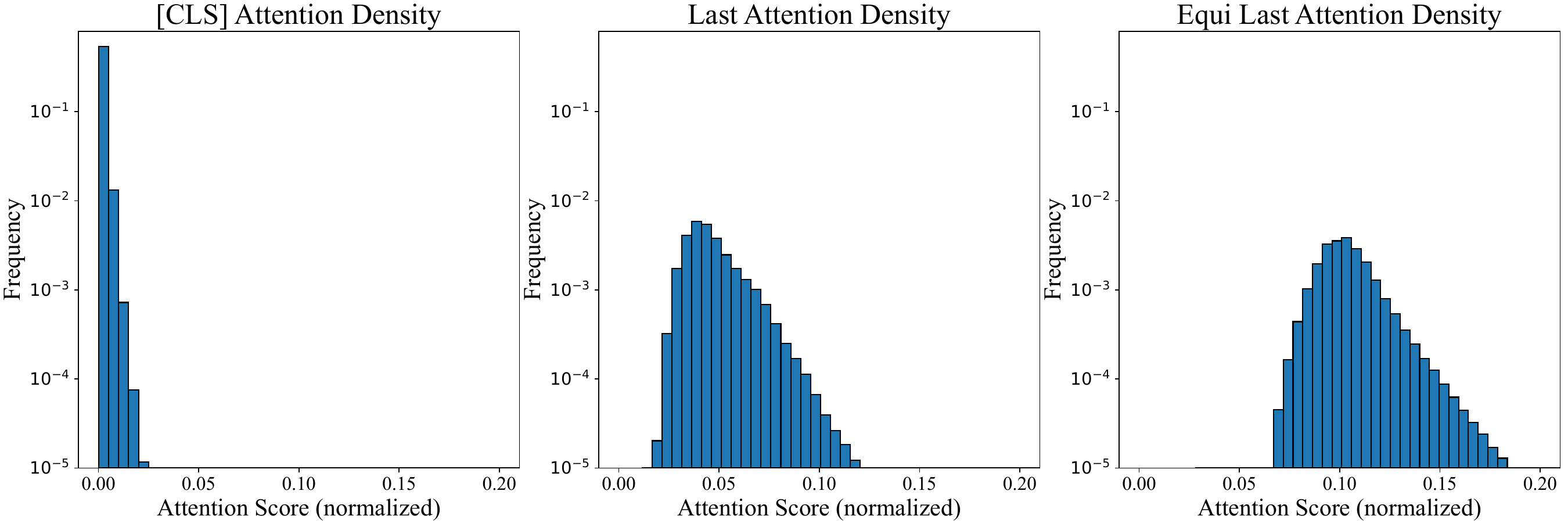}
    \caption{}
    \label{fig:PDF}
  \end{subfigure}

  \begin{subfigure}{1.0\linewidth}
    \includegraphics[width=\linewidth]{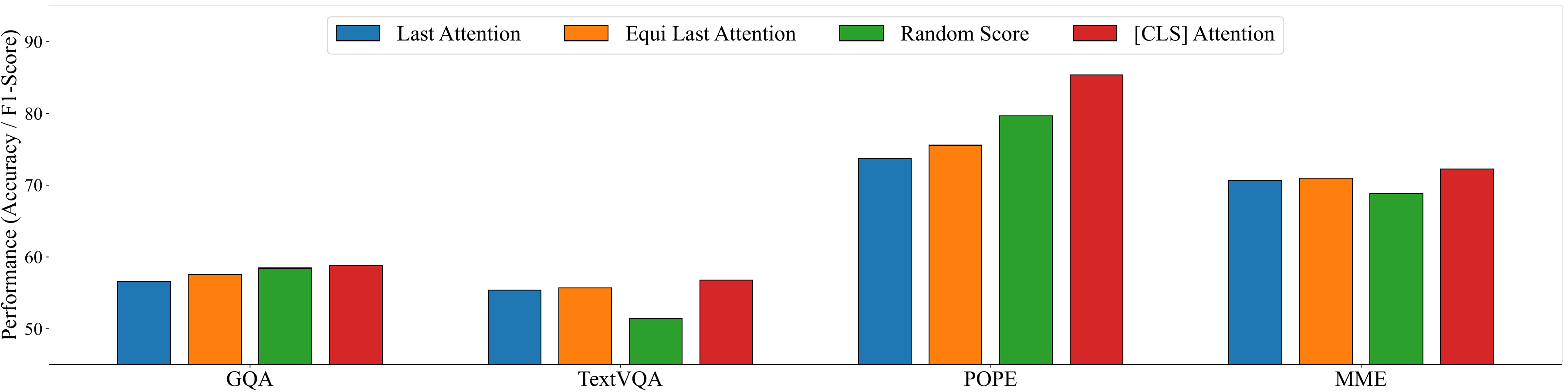}
    \caption{}
    \label{fig:performance_method}
  \end{subfigure}
  \caption{\textbf{Analysis of text-visual attention dispersion.} (a) \textbf{Cumulative distribution of different attentions.} In [\texttt{CLS}] attention, a few visual tokens absorb the majority of attention, while last attention is relatively dispersed. (b) \textbf{Density distribution of different attentions.} Unlike [\texttt{CLS}] attention, the density of last attention, both before and after the elimination of position embedding decay, is obviously high entropy and low peak. (c) \textbf{Performance of pruning based on different attentions across various benchmarks.} Last attention, with position embedding decay removed, achieves performance gains but still underperforms random pruning on some benchmarks due to the dispersion issue, while pruning based on [\texttt{CLS}] attention consistently takes the lead.}
  \label{fig:attn_dispersion}
  \vspace{-5mm}
\end{figure}

Given the negative impact of attention shift caused by rotary position embedding on pruning, an intuitive idea is to use text-visual attention devoid of position embedding decay as a basis for pruning. Unfortunately, although this practice eliminates positional bias, we find another phenomenon present both before and after the removal of position embedding, which we call attention dispersion. Specifically, we refer to \textit{last attention} with identical position embedding for all visual tokens as \textit{equi last attention}, and plot the cumulative distribution curves of different attentions in \cref{fig:CDF}. It is observable that [\texttt{CLS}] attention is more concentrated, with the top 20\% visual tokens absorbing about 80\% attention. In contrast, text-visual attention, whether or not position embedding is removed, is spread across more visual tokens. We also illustrate the density distribution of different attentions in \cref{fig:PDF}. Different from the extreme peak in [\texttt{CLS}] attention, the density of last attention both before and after removing position embedding shows high entropy and low peak, resembling a more uniform distribution. \textbf{Here, highly concentrated attention indicates that the model identifies important tokens with high certainty, whereas a uniform distribution implies greater difficulty in selecting important tokens based on attention scores.}

To verify this, we evaluate the performance of pruning based on different attentions across various benchmarks, as shown in \cref{fig:performance_method}. After eliminating the long-term decay brought by position embedding, pruning results improved across all benchmarks, further demonstrating the adverse impact of positional bias. However, it is worth noting that pruning based on last attention still lags behind random pruning on some benchmarks, especially the hallucination benchmark POPE. We attribute this gap partly to attention dispersion, which makes it difficult to identify important tokens with rich visual information based on attention. Pruning based on [\texttt{CLS}] attention consistently achieves optimal results across all benchmarks, motivating us to exploit visual cues for more effective token pruning in VLMs.
\section{Exploiting Visual Cues for Token Pruning}
\label{sec:method}

\begin{figure*}
  \centering
  \includegraphics[width=\linewidth]{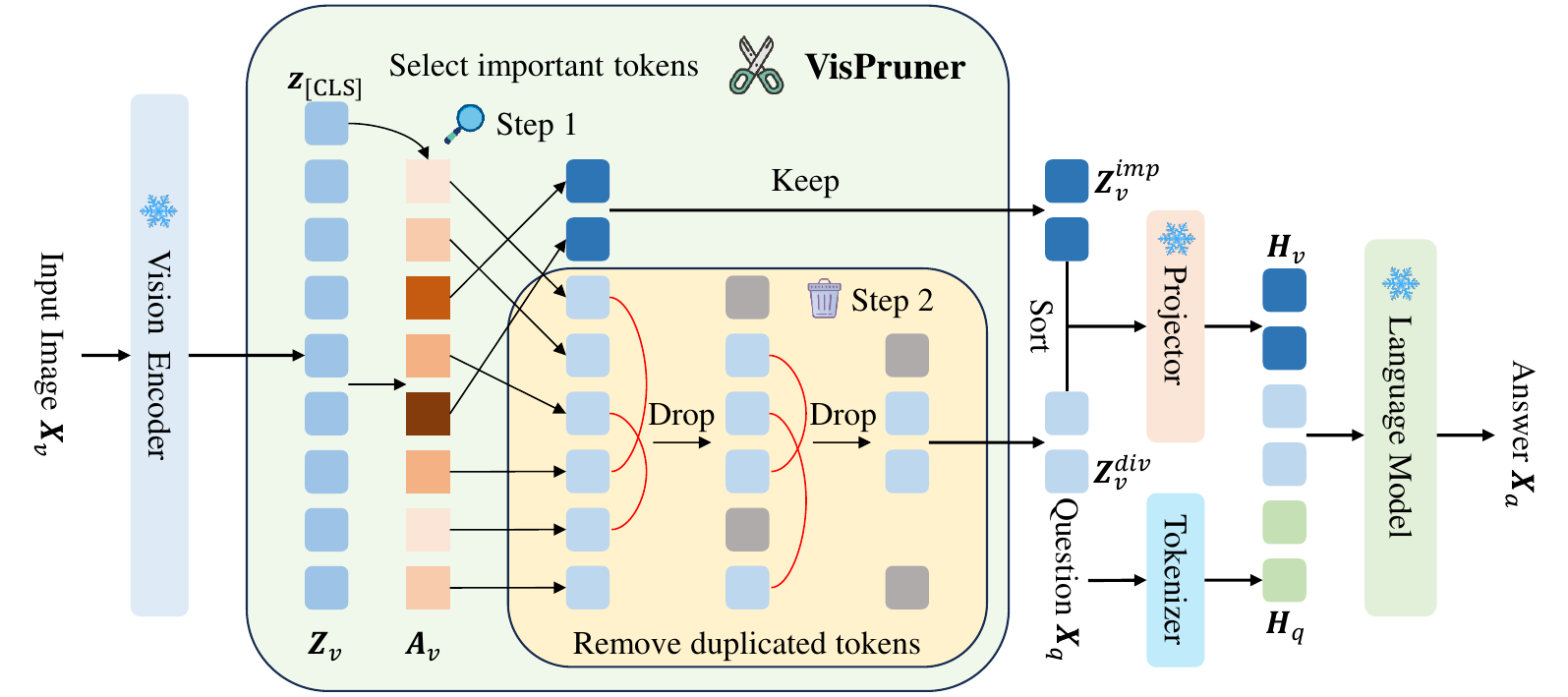}
  \caption{\textbf{Illustration of VisPruner}. We begin by selecting a small portion of important tokens with rich information, based on the [\texttt{CLS}] attention from the visual encoder. For the remaining tokens, we progressively remove duplicates based on similarity, ultimately retaining another set of diverse tokens. These two parts complement each other, ensuring that the model maintains comparable performance even after a significant reduction of visual tokens, without relying on any additional training.}
  \label{fig:pipeline}
  \vspace{-3mm}
\end{figure*}

Based on the above analysis of attention in VLMs, we propose \textbf{VisPruner}, which exploits visual cues for more effective token pruning. We first select a small portion of important tokens with rich information based on the attention from the visual encoder in \cref{sec:important_token}. Then, we retain another set of diverse tokens from the remaining ones based on their similarity as a complement in \cref{sec:diverse_token}. These two parts together maintain the high performance of the model even after a significant reduction of visual tokens. The overall design of our VisPruner is illustrated in \cref{fig:pipeline}.

\subsection{Attention-based important tokens}
\label{sec:important_token}

Visual inputs inherently possess redundancy, and the attention score is a widely used metric to assess token importance and information content~\cite{chen2024fastv, ye2024fitprune, zhang2024sparsevlm, xing2024pyramiddrop}. However, as analyzed before, attention within the language model, due to the presence of shift and dispersion phenomena, does not effectively identify truly important visual tokens. Therefore, in VisPruner, we utilize attention from the visual encoder to select important tokens. Specifically, for the attention matrix $\mA$ calculated in \cref{eq:self-attention}, we first average across the attention heads. Then, for visual encoders like CLIP~\cite{radford2021clip} that contain a [\texttt{CLS}] token, we take the first row of $\mA$ as $\va_v \in \R^{n}$, representing the attention from the [\texttt{CLS}] token to other patch tokens, where $n$ is the length of visual tokens. For models without a [\texttt{CLS}] token (\ie SigLIP~\cite{zhai2023siglip}), we directly average the rows of $\mA$ as the average attention each patch token receives from all other tokens. After getting the visual attention $\va_v$, we can derive a dynamic threshold $\tau$ based on the computation budget $R$:
\begin{equation}
  \tau = \min \left\{ \tau \mid \left| \{ a \in \va_v \mid a \geq \tau \} \right| \leq n \times (1 - R \cdot r) \right\}
\label{eq:threshold}
\end{equation}

Here, $r$ represents the ratio of important tokens within the total computation budget, which is a hyperparameter. In practice, $\tau$ is set as the  $(1 - R \cdot r)$ percentile of $\va_v$. With this threshold, we can select the important part $\mZ_{v}^{imp}$ with rich information from all visual tokens:
\begin{equation}
  \mZ_{v}^{imp} = \{ \vz_v^i \in \mZ_v \mid \va_v^i \geq \tau \}
\label{eq:selection}
\end{equation}

\subsection{Similarity-based diverse tokens}
\label{sec:diverse_token}

Considering that attention in visual encoders often focuses more on foreground objects~\cite{caron2021dino, darcet2023reg}, to prevent performance degradation on background-related questions after pruning, we also retain a set of diverse tokens as a complement. Among the remaining tokens, we first calculate the cosine similarity between each token as a metric of its diversity. Then, we gradually remove tokens with the highest similarity to others until the number of tokens left meets the requirement $R \cdot (1-r)$, implemented in \cref{algo:diverse}.

These diverse tokens $\mZ_{v}^{div}$ better cover more areas of the input image and complement the important tokens, thus preserving as much information in the input image as possible.

\begin{algorithm}
\caption{Similarity-based Duplication Removal}
\label{algo:diverse}
\textbf{Input:} \texttt{normalized}: normalized visual tokens \texttt{(N, C)}, \texttt{n}: diverse token number, \texttt{r}: removal number each iteration
\textbf{Output:} \texttt{remaining\_idx}: index of diverse tokens
\begin{lstlisting}[style=python]
while r > 0:
    remaining = normalized[remaining_idx]

    a, b = remaining[::2], remaining[1::2]
    score = a @ b.transpose(-1, -2)
    score = score.max(dim=-1).values

    diverse_idx = score.argsort(dim=-1, descending=True)[:, r:]
    remaining_idx = cat(remaining_idx[::2][diverse_idx], remaining_idx[1::2], dim=-1)
    r = min(r, remaining_idx.shape[0] - n)
\end{lstlisting}
\vspace{-3mm}
\end{algorithm}

\subsection{Inference with visual pruning}
\label{sec:inference}

With the attention-based important visual tokens $\mZ_{v}^{imp}$ and the similarity-based diverse visual tokens $\mZ_{v}^{div}$, we concatenate these two parts and sort them according to their original positions in the image patch sequence. The merged visual tokens, after passing through a multimodal projector $g$, are then combined with the language instruction $\mH_q$ and fed into the language model $f_\phi$ to generate the response $\mX_a$:
\begin{equation}
  \mH_v = g(\mZ_{v}^{imp} \oplus \mZ_{v}^{div}), \quad \mX_a = f_\phi(\mH_v \oplus \mH_q)
\label{eq:generation}
\end{equation}

Here, $\oplus$ denotes the concatenation operation. Thanks to the complementary design between the important tokens and diverse tokens, our VisPruner is able to maintain considerable performance even at high reduction ratios, significantly enhancing the inference efficiency of VLMs.

\section{Experiments}
\label{sec:experiments}

In this section, we validate our method across various VLM architectures on comprehensive multi-modal tasks, including high-resolution image and video understanding. We compare our approach with multiple existing methods and conduct an ablation study and an efficiency analysis.

\subsection{Experimental setup}

\noindent \textbf{Datasets.} We conduct extensive experiments on 10 image-based multi-modal benchmarks, including common visual question answering tasks such as VQAv2~\cite{goyal2017vqav2}, GQA~\cite{hudson2019gqa}, VizWiz~\cite{gurari2018vizwiz}, ScienceQA-IMG~\cite{lu2022sqa}, and TextVQA~\cite{singh2019textvqa}, as well as other multi-modal benchmarks such as POPE~\cite{li2023pope}, MME~\cite{fu2024mme}, MMBench~\cite{liu2025mmbench}, MMBench-CN~\cite{liu2025mmbench} and MM-Vet~\cite{yu2023mmvet}. Additionally, we also experiment on 4 widely used video question answering benchmarks, including TGIF-QA~\cite{jang2017tgifga}, MSVD-QA~\cite{xu2017msqa}, MSRVTT-QA~\cite{xu2017msqa}, and ActivityNet-QA~\cite{yu2019activitynetqa}. All experiments on these benchmarks follow the default settings and evaluation metrics. Details of each task are provided in the supplementary materials.

\noindent \textbf{Model architectures.} We apply VisPruner to various VLM architectures, including the LLaVA series such as LLaVA-1.5~\cite{liu2024llava1.5}, LLaVA-NeXT~\cite{liu2024llavanext} for high-resolution inputs, and Video-LLaVA~\cite{lin2023videollava} for video understanding, as well as other widely used models like Qwen-VL~\cite{bai2023qwenvl}, InternVL~\cite{chen2024internvl}, and CogVLM~\cite{wang2025cogvlm}. For all models, we follow the same inference settings as the original papers.

\noindent \textbf{Comparison methods.} We choose ToMe~\cite{bolya2022tome}, FastV~\cite{chen2024fastv}, SparseVLM~\cite{zhang2024sparsevlm}, LLaVA-PruMerge~\cite{shang2024llavaprumerge} and the latest VisionZip~\cite{yang2024visionzip} as comparison methods, covering a range of pruning approaches across different stages within VLMs.

\subsection{Main results}

\begin{table*}[t]
  \centering
  \resizebox{\linewidth}{!}{
    \begin{tabular}{l|cccccccccc|cc}
    \toprule
    \textbf{Method} & \textbf{$\text{VQA}^\text{V2}$} & \textbf{GQA} & \textbf{VizWiz} & \textbf{$\text{SQA}^\text{IMG}$} & \textbf{$\text{VQA}^\text{Text}$} & \textbf{POPE} & \textbf{MME} & \textbf{MMB} & \textbf{$\text{MMB}^\text{CN}$} & \textbf{MMVet} & \textbf{Acc.} & \textbf{Rel.} \\
    \rowcolor{lightgray}\multicolumn{13}{c}{\textit{Upper Bound, All 576 Tokens} ($\mathbf{100\%}$)} \\
    LLaVA-1.5-7B & 78.5 & 62.0 & 50.0 & 66.8 & 58.2 & 85.9 & 1510.7 & 64.3 & 58.3 & 31.1 & 63.1 & 100.0\% \\
    \rowcolor{lightgray}\multicolumn{13}{c}{\textit{Retain 128 Tokens} \textcolor{green}{($\downarrow\mathbf{77.8\%}$)}} \\
    ToMe~\cite{bolya2022tome}{\small\texttt{(ICLR23)}} & 63.0 & 52.4 & 50.5 & 59.6 & 49.1 & 62.8 & 1088.4 & 53.3 & 48.8 & 27.2 & 52.1 & 83.9\% \\
    FastV~\cite{chen2024fastv}{\small\texttt{(ECCV24)}} & 61.8 & 49.6 & 51.3 & 60.2 & 50.6 & 59.6 & 1208.9 & 56.1 & 51.4 & 28.1 & 52.9 & 85.4\% \\
    SparseVLM~\cite{zhang2024sparsevlm}{\small\texttt{(ICML25)}} & 73.8 & 56.0 & 51.4 & 67.1 & 54.9 & 80.5 & 1376.2 & 60.0 & 51.1 & 30.0 & 59.4 & 94.4\% \\
    PruMerge+~\cite{shang2024llavaprumerge}{\small\texttt{(2024.05)}} & 74.7 & 57.8 & 52.4 & 67.6 & 54.3 & 81.5 & 1420.5 & 61.3 & 54.7 & 28.7 & 60.4 & 95.8\% \\
    VisionZip~\cite{yang2024visionzip}{\small\texttt{(CVPR25)}} & 75.6 & 57.6 & 52.0 & 68.9 & 56.8 & 83.2 & 1432.4 & 62.0 & 56.7 & 32.6 & 61.7 & 98.4\% \\
    VisPruner{\small\texttt{(Ours)}} & 75.8 & 58.2 & 52.7 & 69.1 & 57.0 & 84.6 & 1461.4 & 62.7 & 57.3 & 33.7 & \textbf{62.4} & \textbf{99.6\%} \\
    \rowcolor{lightgray}\multicolumn{13}{c}{\textit{Retain 64 Tokens}  \textcolor{green}{($\downarrow\mathbf{88.9\%}$)}} \\
    ToMe~\cite{bolya2022tome}{\small\texttt{(ICLR23)}} & 57.1 & 48.6 &  50.2 & 50.0 & 45.3 & 52.5 & 922.3 & 43.7 & 38.9 & 24.1 & 45.7 & 73.9\% \\
    FastV~\cite{chen2024fastv}{\small\texttt{(ECCV24)}} & 55.0 & 46.1 &  50.8 & 51.1 & 47.8 & 48.0 & 1019.6 & 48.0 & 42.7 & 25.8 & 46.6 & 75.9\% \\
    SparseVLM~\cite{zhang2024sparsevlm}{\small\texttt{(ICML25)}} & 68.2 & 52.7 &  50.1 & 62.2 & 51.8 & 75.1 & 1221.1 & 56.2 & 46.1 & 23.3 & 54.7 & 86.4\% \\
    PruMerge+~\cite{shang2024llavaprumerge}{\small\texttt{(2024.05)}} & 67.4 & 54.9 &  52.9 & 68.6 & 53.0 & 77.4 & 1198.2 & 59.3 & 51.0 & 25.9 & 57.0 & 90.6\% \\
    VisionZip~\cite{yang2024visionzip}{\small\texttt{(CVPR25)}} & 72.4 & 55.1 &  52.9 & 69.0 & 55.5 & 77.0 & 1365.6 & 60.1 & 55.4 & 31.7 & 59.7 & 95.6\% \\
    VisPruner{\small\texttt{(Ours)}} & 72.7 & 55.4 & 53.3 & 69.1 & 55.8 & 80.4 & 1369.9 & 61.3 & 55.1 & 32.3 & \textbf{60.4} & \textbf{96.6\%} \\
    \rowcolor{lightgray}\multicolumn{13}{c}{\textit{Retain 32 Tokens}  \textcolor{green}{($\downarrow\mathbf{94.4\%}$)}} \\
    ToMe~\cite{bolya2022tome}{\small\texttt{(ICLR23)}} & 46.8 & 43.6 &  51.3 & 41.4 & 38.3 & 39.0 & 828.4 & 31.6 & 28.1 & 17.3 & 37.9 & 61.4\% \\
    FastV~\cite{chen2024fastv}{\small\texttt{(ECCV24)}} & 43.4 & 41.5 &  51.7 & 42.6 & 42.5 & 32.5 & 884.6 & 37.8 & 33.2 & 20.7 & 39.0 & 64.1\% \\
    SparseVLM~\cite{zhang2024sparsevlm}{\small\texttt{(ICML25)}} & 58.6 & 48.3 &  51.9 & 57.3 & 46.1 & 67.9 & 1046.7 & 51.4 & 40.6 & 18.6 & 49.3 & 77.9\% \\
    PruMerge+~\cite{shang2024llavaprumerge}{\small\texttt{(2024.05)}} & 54.9 & 51.1 &  52.8 & 68.5 & 50.6 & 70.9 & 940.8 & 56.8 & 47.0 & 21.4 & 52.1 & 83.0\% \\
    VisionZip~\cite{yang2024visionzip}{\small\texttt{(CVPR25)}} & 67.1 & 51.8 &  52.9 & 68.8 & 53.1 & 68.7 & 1247.4 & 57.7 & 50.3 & 25.5 & 55.8 & 89.0\% \\
    VisPruner{\small\texttt{(Ours)}} & 67.7 & 52.2 & 53.0 & 69.2 & 53.9 & 72.7 & 1271.0 & 58.4 & 52.7 & 28.8 & \textbf{57.2} & \textbf{91.5\%} \\
    \bottomrule
    \end{tabular}
  }
  \caption{\textbf{Performance comparison of different pruning methods on LLaVA-1.5-7B.} Here, \textbf{Acc.} denotes the average accuracy across 10 benchmarks, and \textbf{Rel.} represents the average percentage of performance maintained at the corresponding reduction ratio.}
  \label{tab:llava-1.5}
  \vspace{-5mm}
\end{table*}

We first apply VisPruner to the classic LLaVA-1.5 model and conduct a comprehensive comparison with existing methods. \cref{tab:llava-1.5} shows the the performance of different pruning methods on the LLaVA-1.5-7B model when retaining only 128, 64, and 32 tokens. With \textbf{77.8\%} of the visual tokens pruned, VisPruner remarkably maintained \textbf{almost all of the original performance}. When the number of retained tokens is further reduced to 64, about one-tenth of the full sequence length, VisPruner only decreases the original performance by \textbf{3.4\%}, outperforming the best text-visual attention-based method, SparseVLM, by as much as \textbf{10.2\%}. It also beats the latest vision-based pruning method, VisionZip, which merges redundant visual tokens in addition to pruning. This result suggests that identifying diverse tokens is more crucial than simply merging similar ones. With only about \textbf{5\%} of the tokens retained, VisPruner is still able to maintain \textbf{91.5\%} of the performance, proving the effectiveness of its design that leverages both important tokens and diverse tokens as complements.

\subsection{VisPruner with higher resolution}

\begin{table}[t]
  \centering
  \resizebox{\linewidth}{!}{
    \begin{tabular}{l|ccccc|cc}
    \toprule
    \textbf{Method} & \textbf{$\text{VQA}^\text{V2}$} & \textbf{GQA} & \textbf{$\text{VQA}^\text{Text}$} & \textbf{POPE} & \textbf{MME} & \textbf{Acc.} & \textbf{Rel.} \\
    \rowcolor{lightgray}\multicolumn{8}{c}{\textit{Upper Bound, All 2880 Tokens} ($\mathbf{100\%}$)} \\
    LLaVA-NeXT-7B & 81.2 & 62.9 &  59.6 & 86.3 & 1513.8 & 73.1 & 100.0\% \\
    \rowcolor{lightgray}\multicolumn{8}{c}{\textit{Retain 640 Tokens} \textcolor{green}{($\downarrow\mathbf{77.8\%}$)}} \\
    FastV~\cite{chen2024fastv}{\small\texttt{(ECCV24)}} & 78.9 & 60.4 &  58.4 & 83.1 & 1477.3 & 70.9 &  97.0\% \\
    SparseVLM~\cite{zhang2024sparsevlm}{\small\texttt{(ICML25)}} & 78.2 & 59.1 &  56.2 & 80.9 & 1456.3 & 69.4 &  94.9\% \\
    VisionZip~\cite{yang2024visionzip}{\small\texttt{(CVPR25)}} & 79.2 & 60.1 &  58.5 & 82.2 & 1468.4 & 70.7 &  96.7\% \\
    VisPruner{\small\texttt{(Ours)}} & 79.8 & 61.6 &  59.3 & 85.9 & 1480.7 & \textbf{72.1} & \textbf{98.6\%} \\
    \rowcolor{lightgray}\multicolumn{8}{c}{\textit{Retain 320 Tokens}  \textcolor{green}{($\downarrow\mathbf{88.9\%}$)}} \\
    FastV~\cite{chen2024fastv}{\small\texttt{(ECCV24)}} & 71.9 & 55.9 &  55.7 & 71.7 & 1282.9 & 63.9 &  87.7\% \\
    SparseVLM~\cite{zhang2024sparsevlm}{\small\texttt{(ICML25)}} & 71.4 & 56.5 &  52.4 & 73.5 & 1342.7 & 64.2 &  87.9\% \\
    VisionZip~\cite{yang2024visionzip}{\small\texttt{(CVPR25)}} & 74.2 & 58.1 &  55.3 & 75.0 & 1348.8 & 66.0 &  90.5\% \\
    VisPruner{\small\texttt{(Ours)}} & 75.7 & 58.4 &  57.6 & 80.4 & 1370.1 & \textbf{68.1} & \textbf{93.3\%} \\
    \rowcolor{lightgray}\multicolumn{8}{c}{\textit{Retain 160 Tokens}  \textcolor{green}{($\downarrow\mathbf{94.4\%}$)}} \\
    FastV~\cite{chen2024fastv}{\small\texttt{(ECCV24)}} & 61.8 & 49.8 &  51.9 & 51.7 & 1079.5 & 53.8 &  74.7\% \\
    SparseVLM~\cite{zhang2024sparsevlm}{\small\texttt{(ICML25)}} & 62.2 & 50.2 &  45.1 & 54.6 & 1167.1 & 54.1 &  74.5\% \\
    VisionZip~\cite{yang2024visionzip}{\small\texttt{(CVPR25)}} & 67.3 & 54.3 &  54.7 & 59.4 & 1239.7 & 59.5 &  82.3\% \\
    VisPruner{\small\texttt{(Ours)}} & 70.6 & 54.7 &  56.0 & 72.9 & 1226.0 & \textbf{63.1} & \textbf{86.7\%} \\
    \bottomrule
    \end{tabular}
  }
  \caption{\textbf{Performance comparison of different pruning methods on LLaVA-NeXT-7B.} \textbf{Acc.} denotes the average accuracy across benchmarks, and \textbf{Rel.} represents the average percentage of performance maintained at the corresponding reduction ratio.}
  \label{tab:llava-next}
  \vspace{-3mm}
\end{table}

Some works~\cite{luo2024llavahr, liu2024llavanext, xu2024llavauhd} attempt to improve VLM performance on visual question answering tasks by increasing the input image resolution. However, this resolution increase introduces much more visual tokens, further intensifying the computational load on the VLM. In this section, we apply VisPruner to LLaVA-NeXT-7B, which can handle up to \textbf{2880} visual tokens. The results are presented in \cref{tab:llava-next}. Compared to LLaVA-1.5, LLaVA-NeXT involves a greater number of visual tokens, implying a higher degree of redundancy. At a \textbf{94.4\%} reduction ratio, retaining only \textbf{160} visual tokens, VisPruner preserves \textbf{86.7\%} of the original performance, significantly outperforming FastV (74.7\%), SparseVLM (74.5\%) as well as VisionZip (82.3\%), demonstrating the value of VisPruner for high-resolution visual inputs.

\subsection{VisPruner with video understanding}

\begin{table}[t]
  \centering
  \resizebox{\linewidth}{!}{
    \begin{tabular}{l|cc|cc|cc|cc}
    \toprule
    \multirow{2}{*}{\textbf{Method}} & \multicolumn{2}{c|}{\textbf{TGIF-QA}} & \multicolumn{2}{c|}{\textbf{MSVD-QA}} & \multicolumn{2}{c|}{\textbf{MSRVTT-QA}} & \multicolumn{2}{c}{\textbf{Average}} \\
     & Acc. & Score & Acc. & Score & Acc. & Score & Acc. & Score \\
    \rowcolor{lightgray}\multicolumn{9}{c}{\textit{Upper Bound, All 2048 Tokens} ($\mathbf{100\%}$)} \\
    Video-LLaVA & 19.8 & 2.53 &  70.5 & 3.93 &  57.5  & 3.50 & 49.3 & 3.32 \\
    \rowcolor{lightgray}\multicolumn{9}{c}{\textit{Retain 455 Tokens} \textcolor{green}{($\downarrow\mathbf{77.8\%}$)}} \\
    FastV~\cite{chen2024fastv}{\small\texttt{(ECCV24)}} & 19.2 & 2.50 &  69.1 & 3.91 &  54.4  & 3.42 & 47.6 & 3.28 \\
    VisPruner{\small\texttt{(Ours)}} & 18.4 & 2.49 &  70.2 & 3.95 &  56.7  & 3.50 & \textbf{48.4} & \textbf{3.31} \\
    \rowcolor{lightgray}\multicolumn{9}{c}{\textit{Retain 227 Tokens}  \textcolor{green}{($\downarrow\mathbf{88.9\%}$)}} \\
    FastV~\cite{chen2024fastv}{\small\texttt{(ECCV24)}} & 14.3 & 2.42 & 68.9 & 3.90 & 53.0 & 3.40 & 45.4 & 3.24 \\
    VisPruner{\small\texttt{(Ours)}} & 15.9 & 2.41 & 69.3 & 3.92 & 55.6 & 3.45 & \textbf{46.9} & \textbf{3.26} \\
    \rowcolor{lightgray}\multicolumn{9}{c}{\textit{Retain 114 Tokens}  \textcolor{green}{($\downarrow\mathbf{94.4\%}$)}} \\
    FastV~\cite{chen2024fastv}{\small\texttt{(ECCV24)}} & 10.6 & 2.29 & 64.1 & 3.78 & 52.4 & 3.39 & 42.4 & 3.15 \\
    VisPruner{\small\texttt{(Ours)}} & 14.1 & 2.35 & 65.4 & 3.79 & 54.1 & 3.41 & \textbf{44.5} & \textbf{3.18} \\
    \bottomrule
    \end{tabular}
  }
  \caption{\textbf{Video understanding performance with Video-LLaVA across three commonly used video question answering benchmarks.} Performance is evaluated using the first 1,000 samples from each benchmark and gpt-3.5-turbo is used for scoring.}
  \label{tab:video-llava}
  \vspace{-3mm}
\end{table}

Video understanding is another scenario with high visual redundancy. We also apply our VisPruner to Video-LLaVA, which accepts videos as input. Following \cite{maaz2023videochatgpt} and \cite{lin2023videollava}, we conduct experiments on three video question answering benchmarks, using ChatGPT-Assistant for evaluation. Due to the commercial API usage limits, we follow \cite{chen2024fastv} to use the first 1K samples from each benchmark in our experiments. The evaluation results are shown in \cref{tab:video-llava}. Video-LLaVA processes 8 frames of 224-resolution video, totaling \textbf{2048} visual tokens. VisPruner maintains 95.1\% of the original performance at a \textbf{88.9\%} reduction ratio (retaining \textbf{227} tokens) and \textbf{90.3\%} with only \textbf{114} tokens (\textbf{94.4\%} reduction ratio), significantly outperforming FastV (86\%). This improvement is due to the higher redundancy in temporally continuous video frames compared to single images, and \textbf{the [\texttt{CLS}] attention accurately identifies key tokens within the video sequence}, enabling VisPruner to maintain strong performance even at high reduction ratios.

\subsection{VisPruner with other VLM architecture}

\begin{table}[t]
  \centering
  \resizebox{\linewidth}{!}{
    \begin{tabular}{l|ccccc|cc}
    \toprule
    \textbf{Method} & \textbf{$\text{VQA}^\text{V2}$} & \textbf{GQA} & \textbf{VizWiz} & \textbf{$\text{SQA}^\text{IMG}$} & \textbf{$\text{VQA}^\text{Text}$} & \textbf{Acc.} & \textbf{Rel.} \\
    \rowcolor{lightgray}\multicolumn{8}{c}{\textit{Upper Bound, All 256 Tokens} ($\mathbf{100\%}$)} \\
    Qwen-VL-7B & 78.8 & 59.3 &  35.2 & 67.1 & 63.8 & 60.8 & 100.0\% \\
    \rowcolor{lightgray}\multicolumn{8}{c}{\textit{Retain 128 Tokens} \textcolor{green}{($\downarrow\mathbf{50\%}$)}} \\
    FastV~\cite{chen2024fastv}{\small\texttt{(ECCV24)}} & 76.5 & 56.9 & 32.7 & 65.3 & 58.2 & 57.9 & 94.9\% \\
    VisPruner{\small\texttt{(Ours)}} & 77.4 & 57.8 & 33.4 & 65.9 & 59.6 & \textbf{58.8} & \textbf{96.4\%} \\
    \rowcolor{lightgray}\multicolumn{8}{c}{\textit{Upper Bound, All 576 Tokens} ($\mathbf{100\%}$)} \\
    InternVL-Chat-13B & 79.3 & 62.9 &  52.6 & 66.3 & 57.0 & 63.6 & 100.0\% \\
    \rowcolor{lightgray}\multicolumn{8}{c}{\textit{Retain 144 Tokens}  \textcolor{green}{($\downarrow\mathbf{75\%}$)}} \\
    FastV~\cite{chen2024fastv}{\small\texttt{(ECCV24)}} & 74.1 & 58.2 & 50.8 & 66.6 & 55.6 & 61.1 & 96.1\% \\
    VisPruner{\small\texttt{(Ours)}} & 76.7 & 60.2 & 51.9 & 67.3 & 55.3 & \textbf{62.3} & \textbf{97.9\%} \\
    \rowcolor{lightgray}\multicolumn{8}{c}{\textit{Upper Bound, All 1225 Tokens} ($\mathbf{100\%}$)} \\
    CogVLM-chat-1.1-17B & 80.9 & 58.2 &  49.6 & 68.4 & 69.2 & 65.3 & 100.0\% \\
    \rowcolor{lightgray}\multicolumn{8}{c}{\textit{Retain 123 Tokens}  \textcolor{green}{($\downarrow\mathbf{90\%}$)}} \\
    FastV~\cite{chen2024fastv}{\small\texttt{(ECCV24)}} & 74.2 & 40.3 & 42.9 & 63.5 & 41.9 & 52.6 & 80.2\% \\
    VisPruner{\small\texttt{(Ours)}} & 74.6 & 48.4 & 46.6 & 68.2 & 59.6 & \textbf{59.5} & \textbf{91.0\%} \\
    \bottomrule
    \end{tabular}
  }
  \caption{\textbf{Performance of VisPruner on different VLM architectures.} \textbf{Acc.} denotes the average accuracy across benchmarks, \textbf{Rel.} represents the average percentage of performance maintained.}
  \label{tab:qwen-internvl-cogvlm}
  \vspace{-3mm}
\end{table}

In addition to the LLaVA series, we also apply VisPruner to other model architectures, including the dominant Qwen-VL~\cite{bai2023qwenvl}, InternVL~\cite{chen2024internvl}, and CogVLM~\cite{wang2025cogvlm}. For Qwen-VL, which uses a Q-Former as a multimodal projector, we first extract attention for each patch token from the visual encoder, and then track each query's attention towards patch tokens in the Q-Former. The integration of these two components yields the visual attention for the final query sequence. InternVL and CogVLM both use MLPs as projectors, so we apply VisPruner to them in the same manner as with LLaVA. The performance at different reduction ratios are shown in \cref{tab:qwen-internvl-cogvlm}. Our VisPruner performs well across a variety of VLM architectures and reduction ratios, consistently outperforming the text-visual attention-based FastV, proving that visual cues can be exploited for effective token pruning across different architectures.

\subsection{Ablation study}

\begin{figure}
  \centering
  \includegraphics[width=\linewidth,height=0.6\linewidth]{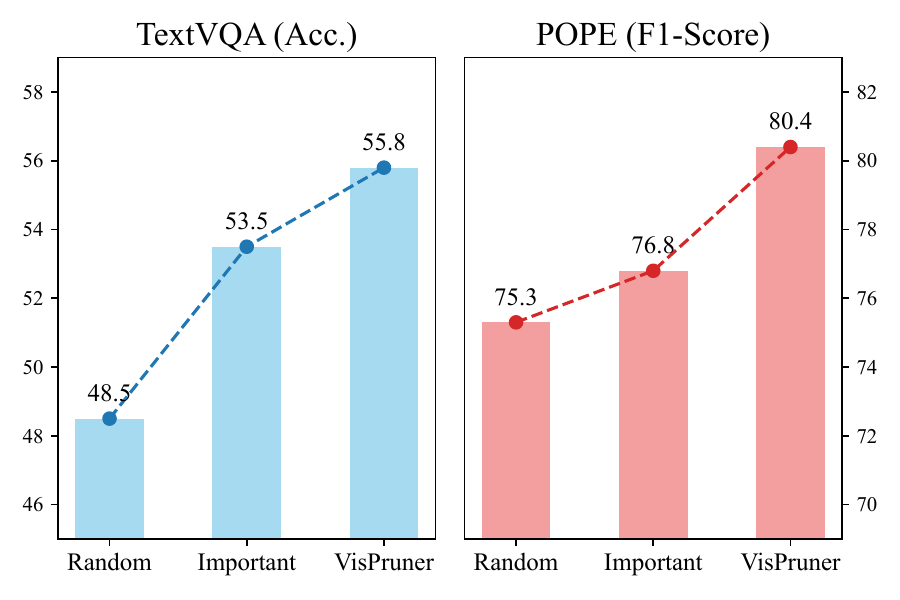}
  \caption{\textbf{Ablation study of the core components.} \textit{Random} denotes randomly selecting tokens, \textit{Important} refers to selecting only tokens with high attention scores, and \textit{VisPruner} represents the final version of our method, which also includes a diverse set of tokens with low redundancy as a complement.}
  \label{fig:ablation}
  \vspace{-5mm}
\end{figure}

To verify the effectiveness of the two core components of our method, we conduct an ablation experiment. We retain 64 visual tokens from different sources in LLaVA-1.5-7B and observe their performance on the TextVQA and POPE benchmarks, as shown in \cref{fig:ablation}. Randomly selecting tokens results in significant performance degradation due to the loss of important visual information. Selecting important tokens through visual attention led to a performance improvement of 5\% and 1.5\% on the two benchmarks. However, as attention in the visual encoder tends to focus on the foreground, retaining only important tokens results in the loss of information related to the background. Here, we also select a set of diverse tokens based on similarity from the remaining as a complement, which led to additional gains of 2.3\% and 3.6\%, respectively, demonstrating the importance of this information complementation for the model.

\subsection{Efficiency analysis}

\begin{table}[t]
  \centering
  \resizebox{\linewidth}{!}{
    \begin{tabular}{l|c|c|c|c|c}
    \toprule
    \textbf{Method}        & \textbf{\# Token}              & \begin{tabular}[c]{@{}c@{}}\textbf{FLOPs}\\ \textbf{(T)}\end{tabular} & \begin{tabular}[c]{@{}c@{}}\textbf{Storage}\\ \textbf{(MB)}\end{tabular} & \begin{tabular}[c]{@{}c@{}}\textbf{GPU Memory}\\ \textbf{(GB)}\end{tabular} & \begin{tabular}[c]{@{}c@{}}\textbf{CUDA Time}\\ \textbf{(ms)}\end{tabular} \\
    \midrule
    LLaVA-NeXT-7B & 2880 & 43.6 & 1440 & 17.0 & 313 \\
    \midrule
    FastV~\cite{chen2024fastv}{\small\texttt{(ECCV24)}} & \multirow{2}{*}{640} & 13.5 & 380 & 16.9 & 148 \\
    VisPruner{\small\texttt{(Ours)}}     &                      & \textbf{11.5} & \textbf{360} & \textbf{14.8} & \textbf{117} \\
    \midrule
    FastV~\cite{chen2024fastv}{\small\texttt{(ECCV24)}} & \multirow{2}{*}{160}  & 6.3 & 95 & 16.9 & 112 \\
    VisPruner{\small\texttt{(Ours)}}     &                       & \textbf{3.8} & \textbf{80} & \textbf{14.7} & \textbf{78} \\
    \bottomrule
    \end{tabular}
  }
  \caption{\textbf{Efficiency analysis with LLaVA-NeXT-7B on POPE.} At the same reduction ratio, VisPruner requires less memory and achieves faster inference compared to FastV.}
  \vspace{-5mm}
  \label{tab:efficiency}
\end{table}

To demonstrate the efficiency of VisPruner, we perform a comparative analysis of FLOPs, cache storage, GPU memory, and CUDA latency with FastV on LLaVA-NeXT-7B.  All experiments are performed on a single NVIDIA A100-80GB GPU, evaluated using the POPE benchmark. As shown in \cref{tab:efficiency}, by retaining only 160 tokens, VisPruner achieves a $\mathbf{\times 4}$ reduction in inference latency for LLaVA-NeXT. Compared to FastV with the same reduction ratio, VisPruner requires less memory and achieves faster inference. Moreover, the design of pruning before the language model enables compatibility with faster attention implementations like FlashAttention~\cite{dao2022flashattention}, which are infeasible for methods like FastV that require access to the text-visual attention information with in the language model.
\section{Discussion}
\label{sec:discussion}

In this discussion, we attempt to explain the outstanding performance of VisPruner. Despite the inherent redundancy in visual inputs, the distribution of visual information within an image is uneven. The semantic information in the foreground region is more dense, while is relatively sparse in the background area. Some concurrent works have also begun to utilize visual information to guide token pruning. ~\cite{wang2024vtccls} leverages visual attention to select key tokens, resulting in the loss of background information. Other studies~\cite{shang2024llavaprumerge, yang2024visionzip} attempt to merge redundant tokens with key tokens, but the fusion process introduces more noise, leading to performance degradation without further training.

We propose a novel method that first uses attention to select important tokens, and then removes redundancy from the remaining ones based on similarity, ultimately retaining a set of diverse tokens as a supplement. This process can be viewed as dynamic pruning across areas with varying information densities; a higher proportion of tokens is retained in foreground areas, while fewer are kept in background areas. Based on this explanation, dynamically partitioning and finely pruning visual tokens according to varying information densities is an interesting direction for future work. 



\clearpage
{
    \small
    \bibliographystyle{ieeenat_fullname}
    \bibliography{main}

\begin{thebibliography}{66}
\providecommand{\natexlab}[1]{#1}
\providecommand{\url}[1]{\texttt{#1}}
\expandafter\ifx\csname urlstyle\endcsname\relax
  \providecommand{\doi}[1]{doi: #1}\else
  \providecommand{\doi}{doi: \begingroup \urlstyle{rm}\Url}\fi

\bibitem[Achiam et~al.(2023)Achiam, Adler, Agarwal, Ahmad, Akkaya, Aleman, Almeida, Altenschmidt, Altman, Anadkat, et~al.]{achiam2023gpt4}
Josh Achiam, Steven Adler, Sandhini Agarwal, Lama Ahmad, Ilge Akkaya, Florencia~Leoni Aleman, Diogo Almeida, Janko Altenschmidt, Sam Altman, Shyamal Anadkat, et~al.
\newblock Gpt-4 technical report.
\newblock \emph{arXiv preprint arXiv:2303.08774}, 2023.

\bibitem[Bai et~al.(2023{\natexlab{a}})Bai, Bai, Chu, Cui, Dang, Deng, Fan, Ge, Han, Huang, et~al.]{bai2023qwen}
Jinze Bai, Shuai Bai, Yunfei Chu, Zeyu Cui, Kai Dang, Xiaodong Deng, Yang Fan, Wenbin Ge, Yu Han, Fei Huang, et~al.
\newblock Qwen technical report.
\newblock \emph{arXiv preprint arXiv:2309.16609}, 2023{\natexlab{a}}.

\bibitem[Bai et~al.(2023{\natexlab{b}})Bai, Bai, Yang, Wang, Tan, Wang, Lin, Zhou, and Zhou]{bai2023qwenvl}
Jinze Bai, Shuai Bai, Shusheng Yang, Shijie Wang, Sinan Tan, Peng Wang, Junyang Lin, Chang Zhou, and Jingren Zhou.
\newblock Qwen-vl: A frontier large vision-language model with versatile abilities.
\newblock \emph{arXiv preprint arXiv:2308.12966}, 2023{\natexlab{b}}.

\bibitem[Bolya et~al.(2022)Bolya, Fu, Dai, Zhang, Feichtenhofer, and Hoffman]{bolya2022tome}
Daniel Bolya, Cheng-Yang Fu, Xiaoliang Dai, Peizhao Zhang, Christoph Feichtenhofer, and Judy Hoffman.
\newblock Token merging: Your vit but faster.
\newblock \emph{arXiv preprint arXiv:2210.09461}, 2022.

\bibitem[Cai et~al.(2024)Cai, Yang, Gao, and Lee]{cai2024m3}
Mu Cai, Jianwei Yang, Jianfeng Gao, and Yong~Jae Lee.
\newblock Matryoshka multimodal models.
\newblock \emph{arXiv preprint arXiv:2405.17430}, 2024.

\bibitem[Caron et~al.(2021)Caron, Touvron, Misra, J{\'e}gou, Mairal, Bojanowski, and Joulin]{caron2021dino}
Mathilde Caron, Hugo Touvron, Ishan Misra, Herv{\'e} J{\'e}gou, Julien Mairal, Piotr Bojanowski, and Armand Joulin.
\newblock Emerging properties in self-supervised vision transformers.
\newblock In \emph{Proceedings of the IEEE/CVF international conference on computer vision}, pages 9650--9660, 2021.

\bibitem[Chen and Dolan(2011)]{chen2011msvd}
David Chen and William~B Dolan.
\newblock Collecting highly parallel data for paraphrase evaluation.
\newblock In \emph{Proceedings of the 49th annual meeting of the association for computational linguistics: human language technologies}, pages 190--200, 2011.

\bibitem[Chen et~al.(2024{\natexlab{a}})Chen, Ye, He, Wang, Khashabi, and Yuille]{chen2024llavolta}
Jieneng Chen, Luoxin Ye, Ju He, Zhao-Yang Wang, Daniel Khashabi, and Alan Yuille.
\newblock Llavolta: Efficient multi-modal models via stage-wise visual context compression.
\newblock \emph{arXiv preprint arXiv:2406.20092}, 2024{\natexlab{a}}.

\bibitem[Chen et~al.(2024{\natexlab{b}})Chen, Zhao, Liu, Bai, Lin, Zhou, and Chang]{chen2024fastv}
Liang Chen, Haozhe Zhao, Tianyu Liu, Shuai Bai, Junyang Lin, Chang Zhou, and Baobao Chang.
\newblock An image is worth 1/2 tokens after layer 2: Plug-and-play inference acceleration for large vision-language models.
\newblock \emph{arXiv preprint arXiv:2403.06764}, 2024{\natexlab{b}}.

\bibitem[Chen et~al.(2024{\natexlab{c}})Chen, Wu, Wang, Su, Chen, Xing, Zhong, Zhang, Zhu, Lu, et~al.]{chen2024internvl}
Zhe Chen, Jiannan Wu, Wenhai Wang, Weijie Su, Guo Chen, Sen Xing, Muyan Zhong, Qinglong Zhang, Xizhou Zhu, Lewei Lu, et~al.
\newblock Internvl: Scaling up vision foundation models and aligning for generic visual-linguistic tasks.
\newblock In \emph{Proceedings of the IEEE/CVF conference on computer vision and pattern recognition}, pages 24185--24198, 2024{\natexlab{c}}.

\bibitem[Dai et~al.(2020)Dai, Lai, Yang, and Le]{dai2020funneltransformer}
Zihang Dai, Guokun Lai, Yiming Yang, and Quoc Le.
\newblock Funnel-transformer: Filtering out sequential redundancy for efficient language processing.
\newblock \emph{Advances in neural information processing systems}, 33:\penalty0 4271--4282, 2020.

\bibitem[Dao et~al.(2022)Dao, Fu, Ermon, Rudra, and R{\'e}]{dao2022flashattention}
Tri Dao, Dan Fu, Stefano Ermon, Atri Rudra, and Christopher R{\'e}.
\newblock Flashattention: Fast and memory-efficient exact attention with io-awareness.
\newblock \emph{Advances in Neural Information Processing Systems}, 35:\penalty0 16344--16359, 2022.

\bibitem[Darcet et~al.(2023)Darcet, Oquab, Mairal, and Bojanowski]{darcet2023reg}
Timoth{\'e}e Darcet, Maxime Oquab, Julien Mairal, and Piotr Bojanowski.
\newblock Vision transformers need registers.
\newblock \emph{arXiv preprint arXiv:2309.16588}, 2023.

\bibitem[Dosovitskiy(2020)]{dosovitskiy2020vit}
Alexey Dosovitskiy.
\newblock An image is worth 16x16 words: Transformers for image recognition at scale.
\newblock \emph{arXiv preprint arXiv:2010.11929}, 2020.

\bibitem[Fu et~al.(2024)Fu, Chen, Shen, Qin, Zhang, Lin, Yang, Zheng, Li, Sun, Wu, and Ji]{fu2024mme}
Chaoyou Fu, Peixian Chen, Yunhang Shen, Yulei Qin, Mengdan Zhang, Xu Lin, Jinrui Yang, Xiawu Zheng, Ke Li, Xing Sun, Yunsheng Wu, and Rongrong Ji.
\newblock Mme: A comprehensive evaluation benchmark for multimodal large language models, 2024.

\bibitem[Goyal et~al.(2017)Goyal, Khot, Summers-Stay, Batra, and Parikh]{goyal2017vqav2}
Yash Goyal, Tejas Khot, Douglas Summers-Stay, Dhruv Batra, and Devi Parikh.
\newblock Making the v in vqa matter: Elevating the role of image understanding in visual question answering.
\newblock In \emph{Proceedings of the IEEE conference on computer vision and pattern recognition}, pages 6904--6913, 2017.

\bibitem[Gurari et~al.(2018)Gurari, Li, Stangl, Guo, Lin, Grauman, Luo, and Bigham]{gurari2018vizwiz}
Danna Gurari, Qing Li, Abigale~J Stangl, Anhong Guo, Chi Lin, Kristen Grauman, Jiebo Luo, and Jeffrey~P Bigham.
\newblock Vizwiz grand challenge: Answering visual questions from blind people.
\newblock In \emph{Proceedings of the IEEE conference on computer vision and pattern recognition}, pages 3608--3617, 2018.

\bibitem[Hong et~al.(2024)Hong, Jiang, Qi, Meng, Yu, Zhou, and Zhou]{hong2024rope}
Xiangyu Hong, Che Jiang, Biqing Qi, Fandong Meng, Mo Yu, Bowen Zhou, and Jie Zhou.
\newblock On the token distance modeling ability of higher rope attention dimension.
\newblock \emph{arXiv preprint arXiv:2410.08703}, 2024.

\bibitem[Hu et~al.(2024)Hu, Dou, Li, Kamath, Peng, and Chang]{hu2024mqt}
Wenbo Hu, Zi-Yi Dou, Liunian~Harold Li, Amita Kamath, Nanyun Peng, and Kai-Wei Chang.
\newblock Matryoshka query transformer for large vision-language models.
\newblock \emph{arXiv preprint arXiv:2405.19315}, 2024.

\bibitem[Huang et~al.(2024)Huang, Dong, Zhang, Wang, He, Wang, Lin, Zhang, and Yu]{huang2024hallucination}
Qidong Huang, Xiaoyi Dong, Pan Zhang, Bin Wang, Conghui He, Jiaqi Wang, Dahua Lin, Weiming Zhang, and Nenghai Yu.
\newblock Opera: Alleviating hallucination in multi-modal large language models via over-trust penalty and retrospection-allocation.
\newblock In \emph{Proceedings of the IEEE/CVF Conference on Computer Vision and Pattern Recognition}, pages 13418--13427, 2024.

\bibitem[Huang et~al.(2022)Huang, Khetan, Bidart, and Karnin]{huang2022pyramidbetr}
Xin Huang, Ashish Khetan, Rene Bidart, and Zohar Karnin.
\newblock Pyramid-bert: Reducing complexity via successive core-set based token selection.
\newblock \emph{arXiv preprint arXiv:2203.14380}, 2022.

\bibitem[Hudson and Manning(2019)]{hudson2019gqa}
Drew~A Hudson and Christopher~D Manning.
\newblock Gqa: A new dataset for real-world visual reasoning and compositional question answering.
\newblock In \emph{Proceedings of the IEEE/CVF conference on computer vision and pattern recognition}, pages 6700--6709, 2019.

\bibitem[Jang et~al.(2017)Jang, Song, Yu, Kim, and Kim]{jang2017tgifga}
Yunseok Jang, Yale Song, Youngjae Yu, Youngjin Kim, and Gunhee Kim.
\newblock Tgif-qa: Toward spatio-temporal reasoning in visual question answering.
\newblock In \emph{Proceedings of the IEEE conference on computer vision and pattern recognition}, pages 2758--2766, 2017.

\bibitem[Jiang et~al.(2023)Jiang, Sablayrolles, Mensch, Bamford, Chaplot, Casas, Bressand, Lengyel, Lample, Saulnier, et~al.]{jiang2023mistral}
Albert~Q Jiang, Alexandre Sablayrolles, Arthur Mensch, Chris Bamford, Devendra~Singh Chaplot, Diego de~las Casas, Florian Bressand, Gianna Lengyel, Guillaume Lample, Lucile Saulnier, et~al.
\newblock Mistral 7b.
\newblock \emph{arXiv preprint arXiv:2310.06825}, 2023.

\bibitem[Krasin et~al.(2017)Krasin, Duerig, Alldrin, Ferrari, Abu-El-Haija, Kuznetsova, Rom, Uijlings, Popov, Veit, et~al.]{krasin2017openimages}
Ivan Krasin, Tom Duerig, Neil Alldrin, Vittorio Ferrari, Sami Abu-El-Haija, Alina Kuznetsova, Hassan Rom, Jasper Uijlings, Stefan Popov, Andreas Veit, et~al.
\newblock Openimages: A public dataset for large-scale multi-label and multi-class image classification.
\newblock \emph{Dataset available from https://github.com/openimages}, 2017.

\bibitem[Krishna et~al.(2017)Krishna, Zhu, Groth, Johnson, Hata, Kravitz, Chen, Kalantidis, Li, Shamma, et~al.]{krishna2017visualgenome}
Ranjay Krishna, Yuke Zhu, Oliver Groth, Justin Johnson, Kenji Hata, Joshua Kravitz, Stephanie Chen, Yannis Kalantidis, Li-Jia Li, David~A Shamma, et~al.
\newblock Visual genome: Connecting language and vision using crowdsourced dense image annotations.
\newblock \emph{International journal of computer vision}, 123:\penalty0 32--73, 2017.

\bibitem[Li et~al.(2023{\natexlab{a}})Li, Li, Savarese, and Hoi]{li2023blip2}
Junnan Li, Dongxu Li, Silvio Savarese, and Steven Hoi.
\newblock Blip-2: Bootstrapping language-image pre-training with frozen image encoders and large language models.
\newblock In \emph{International conference on machine learning}, pages 19730--19742. PMLR, 2023{\natexlab{a}}.

\bibitem[Li et~al.(2024{\natexlab{a}})Li, Yuan, Liu, Tang, Wang, Zhu, and Zhang]{li2024tokenpacker}
Wentong Li, Yuqian Yuan, Jian Liu, Dongqi Tang, Song Wang, Jianke Zhu, and Lei Zhang.
\newblock Tokenpacker: Efficient visual projector for multimodal llm.
\newblock \emph{arXiv preprint arXiv:2407.02392}, 2024{\natexlab{a}}.

\bibitem[Li et~al.(2016)Li, Song, Cao, Tetreault, Goldberg, Jaimes, and Luo]{li2016tgif}
Yuncheng Li, Yale Song, Liangliang Cao, Joel Tetreault, Larry Goldberg, Alejandro Jaimes, and Jiebo Luo.
\newblock Tgif: A new dataset and benchmark on animated gif description.
\newblock In \emph{Proceedings of the IEEE Conference on Computer Vision and Pattern Recognition}, pages 4641--4650, 2016.

\bibitem[Li et~al.(2023{\natexlab{b}})Li, Du, Zhou, Wang, Zhao, and Wen]{li2023pope}
Yifan Li, Yifan Du, Kun Zhou, Jinpeng Wang, Wayne~Xin Zhao, and Ji-Rong Wen.
\newblock Evaluating object hallucination in large vision-language models.
\newblock \emph{arXiv preprint arXiv:2305.10355}, 2023{\natexlab{b}}.

\bibitem[Li et~al.(2024{\natexlab{b}})Li, Wang, and Jia]{li2024llamavid}
Yanwei Li, Chengyao Wang, and Jiaya Jia.
\newblock Llama-vid: An image is worth 2 tokens in large language models.
\newblock In \emph{European Conference on Computer Vision}, pages 323--340. Springer, 2024{\natexlab{b}}.

\bibitem[Lin et~al.(2023)Lin, Ye, Zhu, Cui, Ning, Jin, and Yuan]{lin2023videollava}
Bin Lin, Yang Ye, Bin Zhu, Jiaxi Cui, Munan Ning, Peng Jin, and Li Yuan.
\newblock Video-llava: Learning united visual representation by alignment before projection.
\newblock \emph{arXiv preprint arXiv:2311.10122}, 2023.

\bibitem[Lin et~al.(2014)Lin, Maire, Belongie, Hays, Perona, Ramanan, Doll{\'a}r, and Zitnick]{lin2014mscoco}
Tsung-Yi Lin, Michael Maire, Serge Belongie, James Hays, Pietro Perona, Deva Ramanan, Piotr Doll{\'a}r, and C~Lawrence Zitnick.
\newblock Microsoft coco: Common objects in context.
\newblock In \emph{Computer Vision--ECCV 2014: 13th European Conference, Zurich, Switzerland, September 6-12, 2014, Proceedings, Part V 13}, pages 740--755. Springer, 2014.

\bibitem[Liu et~al.(2024{\natexlab{a}})Liu, Li, Li, and Lee]{liu2024llava1.5}
Haotian Liu, Chunyuan Li, Yuheng Li, and Yong~Jae Lee.
\newblock Improved baselines with visual instruction tuning.
\newblock In \emph{Proceedings of the IEEE/CVF Conference on Computer Vision and Pattern Recognition}, pages 26296--26306, 2024{\natexlab{a}}.

\bibitem[Liu et~al.(2024{\natexlab{b}})Liu, Li, Li, Li, Zhang, Shen, and Lee]{liu2024llavanext}
Haotian Liu, Chunyuan Li, Yuheng Li, Bo Li, Yuanhan Zhang, Sheng Shen, and Yong~Jae Lee.
\newblock Llava-next: Improved reasoning, ocr, and world knowledge, 2024{\natexlab{b}}.

\bibitem[Liu et~al.(2024{\natexlab{c}})Liu, Li, Wu, and Lee]{liu2024llava}
Haotian Liu, Chunyuan Li, Qingyang Wu, and Yong~Jae Lee.
\newblock Visual instruction tuning.
\newblock \emph{Advances in neural information processing systems}, 36, 2024{\natexlab{c}}.

\bibitem[Liu et~al.(2025)Liu, Duan, Zhang, Li, Zhang, Zhao, Yuan, Wang, He, Liu, et~al.]{liu2025mmbench}
Yuan Liu, Haodong Duan, Yuanhan Zhang, Bo Li, Songyang Zhang, Wangbo Zhao, Yike Yuan, Jiaqi Wang, Conghui He, Ziwei Liu, et~al.
\newblock Mmbench: Is your multi-modal model an all-around player?
\newblock In \emph{European Conference on Computer Vision}, pages 216--233. Springer, 2025.

\bibitem[Lu et~al.(2022)Lu, Mishra, Xia, Qiu, Chang, Zhu, Tafjord, Clark, and Kalyan]{lu2022sqa}
Pan Lu, Swaroop Mishra, Tanglin Xia, Liang Qiu, Kai-Wei Chang, Song-Chun Zhu, Oyvind Tafjord, Peter Clark, and Ashwin Kalyan.
\newblock Learn to explain: Multimodal reasoning via thought chains for science question answering.
\newblock \emph{Advances in Neural Information Processing Systems}, 35:\penalty0 2507--2521, 2022.

\bibitem[Luo et~al.(2024)Luo, Zhou, Zhang, Zheng, Sun, and Ji]{luo2024llavahr}
Gen Luo, Yiyi Zhou, Yuxin Zhang, Xiawu Zheng, Xiaoshuai Sun, and Rongrong Ji.
\newblock Feast your eyes: Mixture-of-resolution adaptation for multimodal large language models.
\newblock \emph{arXiv preprint arXiv:2403.03003}, 2024.

\bibitem[Maaz et~al.(2023)Maaz, Rasheed, Khan, and Khan]{maaz2023videochatgpt}
Muhammad Maaz, Hanoona Rasheed, Salman Khan, and Fahad~Shahbaz Khan.
\newblock Video-chatgpt: Towards detailed video understanding via large vision and language models.
\newblock \emph{arXiv preprint arXiv:2306.05424}, 2023.

\bibitem[Nawrot et~al.(2022)Nawrot, Chorowski, {\L}a{\'n}cucki, and Ponti]{nawrot2022dynamicpooling}
Piotr Nawrot, Jan Chorowski, Adrian {\L}a{\'n}cucki, and Edoardo~M Ponti.
\newblock Efficient transformers with dynamic token pooling.
\newblock \emph{arXiv preprint arXiv:2211.09761}, 2022.

\bibitem[Ouyang et~al.(2022)Ouyang, Wu, Jiang, Almeida, Wainwright, Mishkin, Zhang, Agarwal, Slama, Ray, et~al.]{ouyang2022instructgpt}
Long Ouyang, Jeffrey Wu, Xu Jiang, Diogo Almeida, Carroll Wainwright, Pamela Mishkin, Chong Zhang, Sandhini Agarwal, Katarina Slama, Alex Ray, et~al.
\newblock Training language models to follow instructions with human feedback.
\newblock \emph{Advances in neural information processing systems}, 35:\penalty0 27730--27744, 2022.

\bibitem[Radford et~al.(2021)Radford, Kim, Hallacy, Ramesh, Goh, Agarwal, Sastry, Askell, Mishkin, Clark, et~al.]{radford2021clip}
Alec Radford, Jong~Wook Kim, Chris Hallacy, Aditya Ramesh, Gabriel Goh, Sandhini Agarwal, Girish Sastry, Amanda Askell, Pamela Mishkin, Jack Clark, et~al.
\newblock Learning transferable visual models from natural language supervision.
\newblock In \emph{International conference on machine learning}, pages 8748--8763. PMLR, 2021.

\bibitem[Rae et~al.(2019)Rae, Potapenko, Jayakumar, and Lillicrap]{rae2019compressivetransformer}
Jack~W Rae, Anna Potapenko, Siddhant~M Jayakumar, and Timothy~P Lillicrap.
\newblock Compressive transformers for long-range sequence modelling.
\newblock \emph{arXiv preprint arXiv:1911.05507}, 2019.

\bibitem[Shang et~al.(2024)Shang, Cai, Xu, Lee, and Yan]{shang2024llavaprumerge}
Yuzhang Shang, Mu Cai, Bingxin Xu, Yong~Jae Lee, and Yan Yan.
\newblock Llava-prumerge: Adaptive token reduction for efficient large multimodal models.
\newblock \emph{arXiv preprint arXiv:2403.15388}, 2024.

\bibitem[Singh et~al.(2019)Singh, Natarajan, Shah, Jiang, Chen, Batra, Parikh, and Rohrbach]{singh2019textvqa}
Amanpreet Singh, Vivek Natarajan, Meet Shah, Yu Jiang, Xinlei Chen, Dhruv Batra, Devi Parikh, and Marcus Rohrbach.
\newblock Towards vqa models that can read.
\newblock In \emph{Proceedings of the IEEE/CVF conference on computer vision and pattern recognition}, pages 8317--8326, 2019.

\bibitem[Su et~al.(2024)Su, Ahmed, Lu, Pan, Bo, and Liu]{su2024roformer}
Jianlin Su, Murtadha Ahmed, Yu Lu, Shengfeng Pan, Wen Bo, and Yunfeng Liu.
\newblock Roformer: Enhanced transformer with rotary position embedding.
\newblock \emph{Neurocomputing}, 568:\penalty0 127063, 2024.

\bibitem[Team et~al.(2023)Team, Anil, Borgeaud, Alayrac, Yu, Soricut, Schalkwyk, Dai, Hauth, Millican, et~al.]{team2023gemini}
Gemini Team, Rohan Anil, Sebastian Borgeaud, Jean-Baptiste Alayrac, Jiahui Yu, Radu Soricut, Johan Schalkwyk, Andrew~M Dai, Anja Hauth, Katie Millican, et~al.
\newblock Gemini: a family of highly capable multimodal models.
\newblock \emph{arXiv preprint arXiv:2312.11805}, 2023.

\bibitem[Tong et~al.(2024)Tong, Liu, Zhai, Ma, LeCun, and Xie]{tong2024shortcoming}
Shengbang Tong, Zhuang Liu, Yuexiang Zhai, Yi Ma, Yann LeCun, and Saining Xie.
\newblock Eyes wide shut? exploring the visual shortcomings of multimodal llms.
\newblock In \emph{Proceedings of the IEEE/CVF Conference on Computer Vision and Pattern Recognition}, pages 9568--9578, 2024.

\bibitem[Touvron et~al.(2023{\natexlab{a}})Touvron, Lavril, Izacard, Martinet, Lachaux, Lacroix, Rozi{\`e}re, Goyal, Hambro, Azhar, et~al.]{touvron2023llama}
Hugo Touvron, Thibaut Lavril, Gautier Izacard, Xavier Martinet, Marie-Anne Lachaux, Timoth{\'e}e Lacroix, Baptiste Rozi{\`e}re, Naman Goyal, Eric Hambro, Faisal Azhar, et~al.
\newblock Llama: Open and efficient foundation language models.
\newblock \emph{arXiv preprint arXiv:2302.13971}, 2023{\natexlab{a}}.

\bibitem[Touvron et~al.(2023{\natexlab{b}})Touvron, Martin, Stone, Albert, Almahairi, Babaei, Bashlykov, Batra, Bhargava, Bhosale, et~al.]{touvron2023llama2}
Hugo Touvron, Louis Martin, Kevin Stone, Peter Albert, Amjad Almahairi, Yasmine Babaei, Nikolay Bashlykov, Soumya Batra, Prajjwal Bhargava, Shruti Bhosale, et~al.
\newblock Llama 2: Open foundation and fine-tuned chat models.
\newblock \emph{arXiv preprint arXiv:2307.09288}, 2023{\natexlab{b}}.

\bibitem[Vaswani(2017)]{vaswani2017transformer}
A Vaswani.
\newblock Attention is all you need.
\newblock \emph{Advances in Neural Information Processing Systems}, 2017.

\bibitem[Wang et~al.(2024)Wang, Sun, Chen, Lin, Han, and Ding]{wang2024vtccls}
Ao Wang, Fengyuan Sun, Hui Chen, Zijia Lin, Jungong Han, and Guiguang Ding.
\newblock [cls] token tells everything needed for training-free efficient mllms.
\newblock \emph{arXiv preprint arXiv:2412.05819}, 2024.

\bibitem[Wang et~al.(2025)Wang, Lv, Yu, Hong, Qi, Wang, Ji, Yang, Zhao, XiXuan, et~al.]{wang2025cogvlm}
Weihan Wang, Qingsong Lv, Wenmeng Yu, Wenyi Hong, Ji Qi, Yan Wang, Junhui Ji, Zhuoyi Yang, Lei Zhao, Song XiXuan, et~al.
\newblock Cogvlm: Visual expert for pretrained language models.
\newblock \emph{Advances in Neural Information Processing Systems}, 37:\penalty0 121475--121499, 2025.

\bibitem[Xing et~al.(2024)Xing, Huang, Dong, Lu, Zhang, Zang, Cao, He, Wang, Wu, et~al.]{xing2024pyramiddrop}
Long Xing, Qidong Huang, Xiaoyi Dong, Jiajie Lu, Pan Zhang, Yuhang Zang, Yuhang Cao, Conghui He, Jiaqi Wang, Feng Wu, et~al.
\newblock Pyramiddrop: Accelerating your large vision-language models via pyramid visual redundancy reduction.
\newblock \emph{arXiv preprint arXiv:2410.17247}, 2024.

\bibitem[Xu et~al.(2017)Xu, Zhao, Xiao, Wu, Zhang, He, and Zhuang]{xu2017msqa}
Dejing Xu, Zhou Zhao, Jun Xiao, Fei Wu, Hanwang Zhang, Xiangnan He, and Yueting Zhuang.
\newblock Video question answering via gradually refined attention over appearance and motion.
\newblock In \emph{Proceedings of the 25th ACM international conference on Multimedia}, pages 1645--1653, 2017.

\bibitem[Xu et~al.(2016)Xu, Mei, Yao, and Rui]{xu2016msrvtt}
Jun Xu, Tao Mei, Ting Yao, and Yong Rui.
\newblock Msr-vtt: A large video description dataset for bridging video and language.
\newblock In \emph{Proceedings of the IEEE conference on computer vision and pattern recognition}, pages 5288--5296, 2016.

\bibitem[Xu et~al.(2024)Xu, Yao, Guo, Cui, Ni, Ge, Chua, Liu, Sun, and Huang]{xu2024llavauhd}
Ruyi Xu, Yuan Yao, Zonghao Guo, Junbo Cui, Zanlin Ni, Chunjiang Ge, Tat-Seng Chua, Zhiyuan Liu, Maosong Sun, and Gao Huang.
\newblock Llava-uhd: an lmm perceiving any aspect ratio and high-resolution images.
\newblock \emph{arXiv preprint arXiv:2403.11703}, 2024.

\bibitem[Yang et~al.(2024)Yang, Chen, Tian, Wang, Li, Yu, and Jia]{yang2024visionzip}
Senqiao Yang, Yukang Chen, Zhuotao Tian, Chengyao Wang, Jingyao Li, Bei Yu, and Jiaya Jia.
\newblock Visionzip: Longer is better but not necessary in vision language models.
\newblock \emph{arXiv preprint arXiv:2412.04467}, 2024.

\bibitem[Ye et~al.(2024)Ye, Wu, Lin, and Zhou]{ye2024fitprune}
Weihao Ye, Qiong Wu, Wenhao Lin, and Yiyi Zhou.
\newblock Fit and prune: Fast and training-free visual token pruning for multi-modal large language models.
\newblock \emph{arXiv preprint arXiv:2409.10197}, 2024.

\bibitem[Yu et~al.(2023)Yu, Yang, Li, Wang, Lin, Liu, Wang, and Wang]{yu2023mmvet}
Weihao Yu, Zhengyuan Yang, Linjie Li, Jianfeng Wang, Kevin Lin, Zicheng Liu, Xinchao Wang, and Lijuan Wang.
\newblock Mm-vet: Evaluating large multimodal models for integrated capabilities.
\newblock \emph{arXiv preprint arXiv:2308.02490}, 2023.

\bibitem[Yu et~al.(2019)Yu, Xu, Yu, Yu, Zhao, Zhuang, and Tao]{yu2019activitynetqa}
Zhou Yu, Dejing Xu, Jun Yu, Ting Yu, Zhou Zhao, Yueting Zhuang, and Dacheng Tao.
\newblock Activitynet-qa: A dataset for understanding complex web videos via question answering.
\newblock In \emph{Proceedings of the AAAI Conference on Artificial Intelligence}, pages 9127--9134, 2019.

\bibitem[Zhai et~al.(2023)Zhai, Mustafa, Kolesnikov, and Beyer]{zhai2023siglip}
Xiaohua Zhai, Basil Mustafa, Alexander Kolesnikov, and Lucas Beyer.
\newblock Sigmoid loss for language image pre-training.
\newblock In \emph{Proceedings of the IEEE/CVF international conference on computer vision}, pages 11975--11986, 2023.

\bibitem[Zhang et~al.(2025)Zhang, Fang, Yang, and Feng]{zhang2025llavamini}
Shaolei Zhang, Qingkai Fang, Zhe Yang, and Yang Feng.
\newblock Llava-mini: Efficient image and video large multimodal models with one vision token.
\newblock \emph{arXiv preprint arXiv:2501.03895}, 2025.

\bibitem[Zhang et~al.(2024)Zhang, Fan, Ma, Zheng, Huang, Cheng, Gudovskiy, Okuno, Nakata, Keutzer, et~al.]{zhang2024sparsevlm}
Yuan Zhang, Chun-Kai Fan, Junpeng Ma, Wenzhao Zheng, Tao Huang, Kuan Cheng, Denis Gudovskiy, Tomoyuki Okuno, Yohei Nakata, Kurt Keutzer, et~al.
\newblock Sparsevlm: Visual token sparsification for efficient vision-language model inference.
\newblock \emph{arXiv preprint arXiv:2410.04417}, 2024.

\bibitem[Zhao et~al.(2024)Zhao, Han, Tang, Li, Song, Wang, Wang, and You]{zhao2024sgl}
Wangbo Zhao, Yizeng Han, Jiasheng Tang, Zhikai Li, Yibing Song, Kai Wang, Zhangyang Wang, and Yang You.
\newblock A stitch in time saves nine: Small vlm is a precise guidance for accelerating large vlms.
\newblock \emph{arXiv preprint arXiv:2412.03324}, 2024.

\end{thebibliography}
}

\clearpage
\setcounter{page}{1}
\maketitlesupplementary

\section{Details of experimental setup}
\label{sec:detail}

\subsection{Datasets}
\label{sec:detail_dataset}

We evaluate our method on a total of 13 widely used benchmarks, including 10 image benchmarks and 3 video benchmarks. Each task is described as follows.

\subsubsection{Image benchmarks}

We conduct experiments on 10 image benchmarks used in LLaVA~\cite{liu2024llava}, including 5 visual question answering benchmarks and 5 multi-modal reasoning benchmarks. All inference settings and evaluation metrics for these tasks follow the original configurations in LLaVA-1.5~\cite{liu2024llava}.

\noindent \textbf{VQAv2~\cite{goyal2017vqav2}.} The VQAv2 benchmark evaluates the model's visual recognition capabilities through open-ended questions. It consists of 265,016 images from MSCOCO dataset~\cite{lin2014mscoco}, with each image containing at least 3 questions. The dataset incorporates adversarially balanced question design, ensuring that each question corresponds to at least two images with completely different answers, preventing models from relying solely on statistical patterns to derive answers. We utilize the test-dev set for evaluation, which includes 107,394 image-question pairs. Each question is associated with 10 ground truth answers, and automatic evaluation metrics are used for scoring.

\noindent \textbf{GQA~\cite{hudson2019gqa}.} The GQA benchmark focuses on evaluating structured understanding and reasoning abilities for scenes depicted in images. In addition to images and questions, it provides scene graph annotations derived from the Visual Genome dataset~\cite{krishna2017visualgenome} for each image, which include structured descriptions of objects, attributes, and their relationships within the scene. The questions are generated using the scene graphs and a pre-designed engine, ensuring that each question corresponds to a clear semantic path. We use the accuracy on the test-dev set for evaluation, which contains 12,578 image-question pairs.

\noindent \textbf{VizWiz~\cite{gurari2018vizwiz}.} The VizWiz benchmark uses images captured by blind users to evaluate the model's visual understanding capabilities in real-world scenarios. Each image is first taken and uploaded by a blind user, accompanied by a question. The question is then paired with 10 crowdsourced answers for automated evaluation. Since the images are captured by blind users in real-life settings, some questions may be difficult to answer due to issues like blur or poor lighting. Additionally, since the images and questions originate from the same source, some questions may not be directly relevant to the image. We evaluate the model using the test-dev set, which includes 8,000 image-question pairs.

\noindent \textbf{ScienceQA~\cite{lu2022sqa}.} The ScienceQA benchmark uses multiple-choice questions to evaluate the model's zero-shot generalization on scientific topics. The dataset contains rich domain diversity across three subjects: natural sciences, language science, and social science. Questions within each subject are hierarchically organized by topic, category, and skill, encompassing a total of 26 topics, 127 categories, and 379 skills. The images are illustrations related to the questions, and some questions do not have corresponding images. We evaluate the model using a subset of the test set that includes both questions and images, referred to as SQA-IMG, which contains 2,017 image-question pairs.

\noindent \textbf{TextVQA~\cite{singh2019textvqa}.} The TextVQA benchmark is designed to evaluate the model's ability to recognize textual information within images, emphasizing the integration of optical character recognition (OCR) and natural language understanding. The images are primarily sourced from the Open Images v3 dataset~\cite{krasin2017openimages} and contain a variety of scenarios such as signs, billboards, and product packaging that contain rich text information. In addition to raw images, reference OCR tokens are also provided. Answers to the questions may be directly derived from the text in the images or require contextual reasoning. We evaluate the model's performance on a validation set containing 5,000 image-question pairs.

\noindent \textbf{POPE~\cite{li2023pope}.} The POPE benchmark evaluates the hallucination in large vision-language models through questions about object presence. The images are sourced from the MSCOCO dataset~\cite{lin2014mscoco}, and the questions focus on whether a specific object is present in the image, assessing the degree of object hallucination. We use the average F1 score across three different sampling strategies in the test set for evaluation, including 8,910 image-question pairs.

\noindent \textbf{MME~\cite{fu2024mme}.} The MME benchmark aims to comprehensively evaluate the perceptual and cognitive capabilities of multi-modal models, encompassing a total of 14 subtasks. The perception tasks include OCR as well as coarse- and fine-grained recognition. Coarse-grained recognition primarily focuses on the presence, count, position, and color of objects, while fine-grained recognition involves identifying specific posters, celebrities, scenes, landmarks, and artworks. All questions are binary judgment tasks. We use the perception score for performance evaluation, with 2.374 image-question pairs in total.

\noindent \textbf{MMBench~\cite{liu2025mmbench}.} The MMBench benchmark is designed to comprehensively evaluate the capabilities of multi-modal models. It defines three levels of competence from the top down, with the first level containing two basic abilities, perception and reasoning, the second level containing 6 more specific capabilities, and the third level containing 20 concrete tasks. Each task contains multiple choice questions to assess model performance on the task. The benchmark is available in both English and Chinese. The English version includes 4,377 image-question pairs, while the Chinese version, also referred to as \textbf{MMBench-CN}, contains 4,329 pairs. Both versions are used for evaluation.

\noindent \textbf{MM-Vet~\cite{yu2023mmvet}.} The MM-Vet benchmark focuses on the integration of different core vision-language capabilities. It defines 6 core capabilities, including recognition, OCR, knowledge, language generation, spatial awareness, and mathematics, which are combined into 16 specific tasks. The benchmark utilizes ChatGPT assistant for evaluation, providing unified metrics for assessing answers of varying styles. It includes a total of 218 image-question pairs.

\subsubsection{Video benchmarks}

To evaluate the performance of different methods in scenarios with higher visual redundancy, we also conduct experiments on 4 video benchmarks used in Video-LLaVA~\cite{lin2023videollava}. The evaluation follows Video-ChatGPT~\cite{maaz2023videochatgpt}, using \textit{gpt-3.5-turbo} assistant for scoring. Due to the commercial API usage limits, we follow \cite{chen2024fastv} to use the first 1K samples of each benchmark in the experiments.

\noindent \textbf{TGIF-QA~\cite{jang2017tgifga}.} The TGIF-QA benchmark extends image-based VQA tasks to videos, requiring models to focus on both spatial and temporal attentions. It includes 72K animated GIFs from the Tumblr GIF dataset~\cite{li2016tgif} and 165K crowdsourced question-answer pairs. We evaluate model performance using the Frame QA task in this benchmark.

\noindent \textbf{MSVD-QA~\cite{xu2017msqa}.} The MSVD-QA benchmark is based on the Microsoft Research Video Description Corpus~\cite{chen2011msvd}, which is commonly used for video captioning tasks. The question-answer pairs in the benchmark are derived from the descriptions in the corpus. The benchmark consists of 1,970 video clips and 50.5K question-answer pairs in total.

\noindent \textbf{MSRVTT-QA~\cite{xu2017msqa}.} The MSRVTT-QA benchmark is based on the Microsoft Research Video to Text dataset~\cite{xu2016msrvtt}, which is larger and has more complex scenes than the MSVD dataset. The benchmark consists of 10K video clips and 243K question-answer pairs in total.

\subsection{Model architectures}

\noindent \textbf{LLaVA-1.5~\cite{liu2024llava1.5}.} LLaVA is one of the most widely used open-source vision-language models, and its simple design, low tuning cost, and outstanding performance make it a cornerstone in the field of multi-modal models. Specifically, LLaVA employs a pre-trained CLIP as the visual encoder and Vicuna as the text decoder. A simple linear projector connects the two modules, enabling the LLM to accept visual tokens of CLIP as input. Meanwhile, visual instruction tuning allows the model to handle vision-language tasks. Compared to the original LLaVA, LLaVA-1.5 increases the input image resolution from 224 to 336 and incorporates more instruction tuning data, resulting in a significant performance improvement.

\noindent \textbf{LLaVA-NeXT~\cite{liu2024llavanext}.} Also known as LLaVA-1.6, LLaVA-NeXT builds upon LLaVA-1.5 by further increasing the input image resolution, achieving improvements in reasoning, OCR, and world knowledge. Unlike the fixed resolution increase in LLaVA-1.5, LLaVA-NeXT employs a dynamic high-resolution design. Specifically, the model can select the best aspect ratio based on the resolution of the input image, increasing the resolution by up to 4×. Without altering the visual encoder, high-resolution images are split into several sub-images of the same size as the original image. These sub-images are individually encoded and concatenated before being fed into the LLM.

\noindent \textbf{Video-LLaVA~\cite{lin2023videollava}.} On the basis of image understanding, Video-LLaVA extends this capability to video comprehension. It unifies representations of images and videos through alignment before projection. The overall architecture remains consistent with LLaVA: the visual encoder encodes continuous video frames individually, and the representations are concatenated as inputs to the LLM. After joint training, Video-LLaVA is capable of understanding both image and video data.

\noindent \textbf{Qwen-VL~\cite{bai2023qwenvl}.} Qwen-VL is another widely used open-source vision-language model. Similar to LLaVA, it includes a visual encoder (OpenCLIP) and a text decoder (Qwen LLM). For the vision-text connector, Qwen-VL employs a vision-language adapter, which transforms image inputs into fixed-length token sequences via cross-attention. After three stages of training, Qwen-VL achieves strong vision-language understanding capabilities. And Qwen-VL-Chat is further fine-tuned based on Qwen-VL to enhance its performance in conversational tasks.

\begin{figure*}[t]
  \centering
  
  \begin{subfigure}{0.24\linewidth}
    \includegraphics[width=\linewidth]{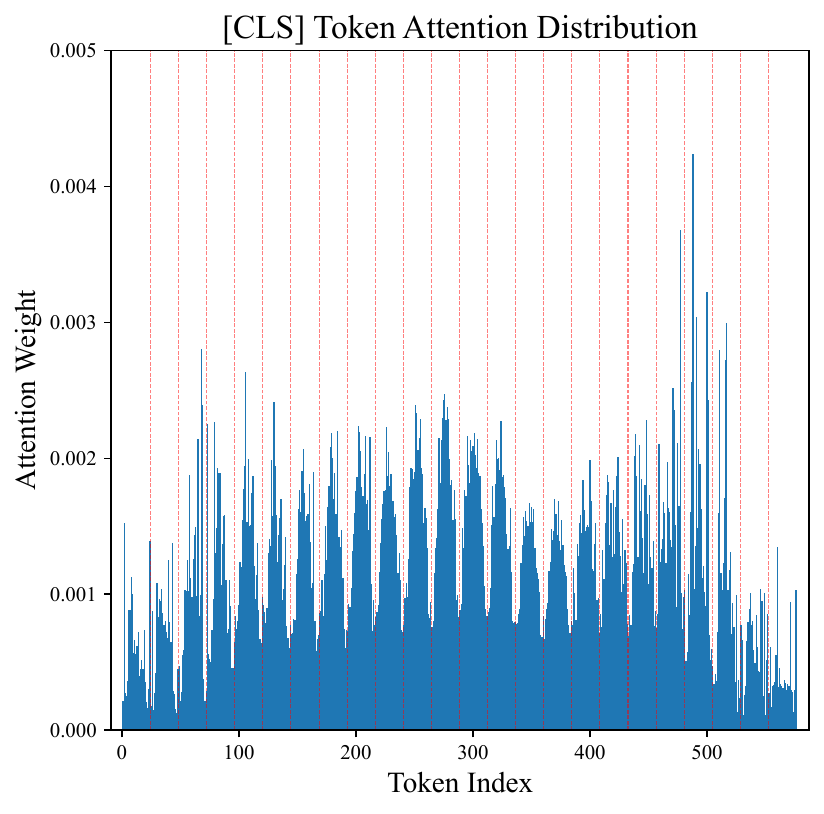}
    \label{fig:cls_attn_penu}
  \end{subfigure}
  \hfill 
  \begin{subfigure}{0.24\linewidth}
    \includegraphics[width=\linewidth]{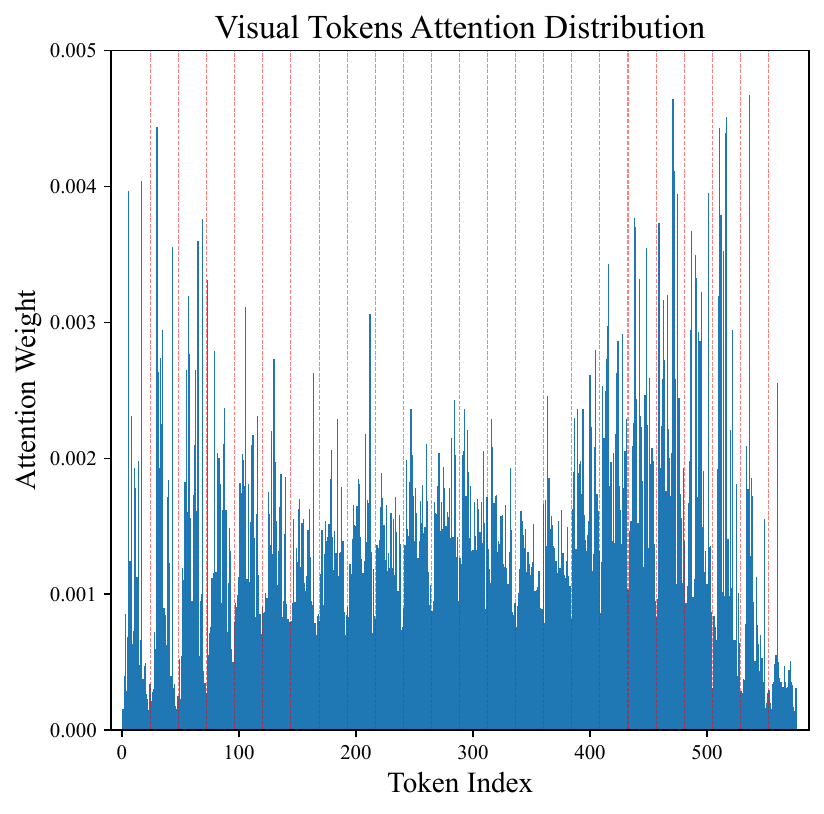}
    \label{fig:vis_attn_penu}
  \end{subfigure}
  \hfill
  \begin{subfigure}{0.24\linewidth}
    \includegraphics[width=\linewidth]{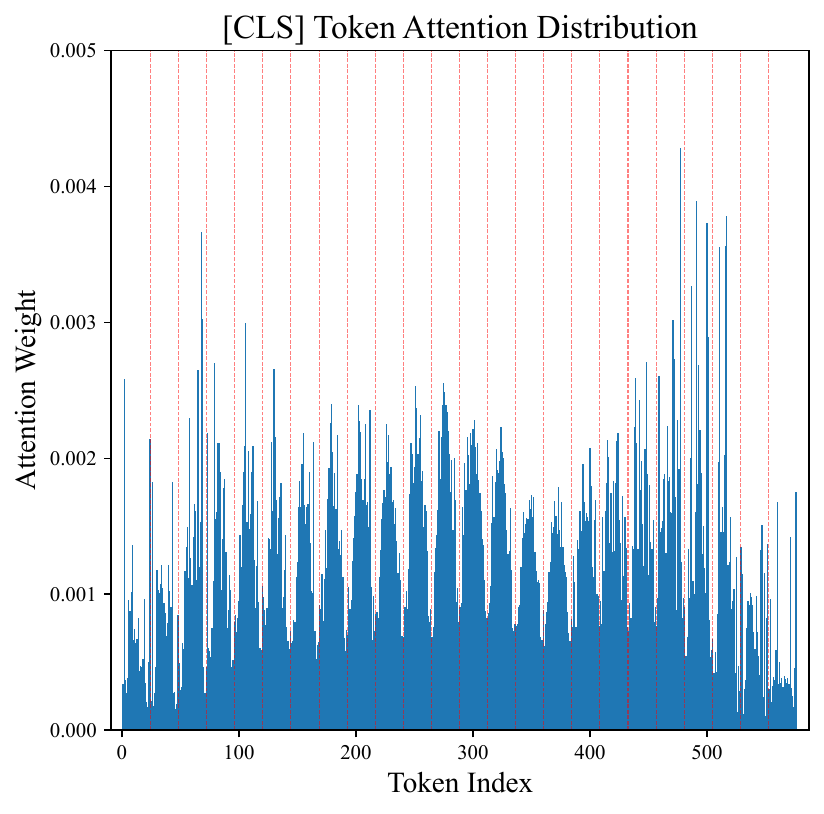}
    \label{fig:cls_attn_last}
  \end{subfigure}
  \hfill
  \begin{subfigure}{0.24\linewidth}
    \includegraphics[width=\linewidth]{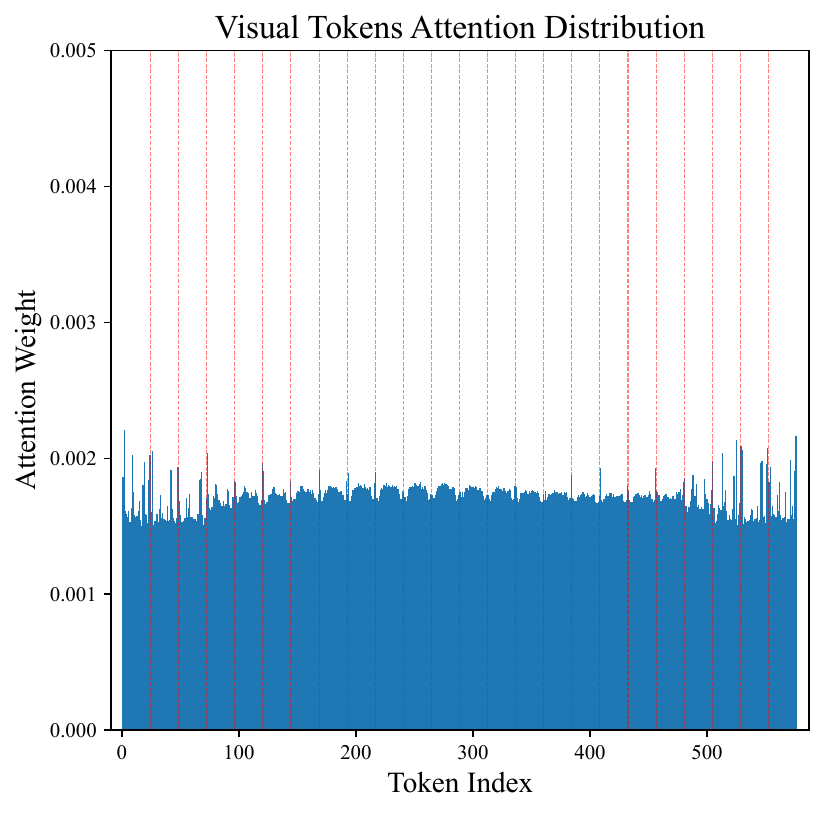}
    \label{fig:vis_attn_last}
  \end{subfigure}
  
  \caption{Distribution of visual attention over token positions in CLIP. From left to right: attention distribution of the [\texttt{CLS}] token in the penultimate layer, average attention distribution across all visual tokens in the penultimate layer, attention distribution of the [\texttt{CLS}] token in the final layer, and average attention distribution across all visual tokens in the final layer. The red vertical dashed lines indicate the length of each row in the input image (24 for CLIP-ViT-L-14-336px used in LLaVA-1.5).}
  \label{fig:clip_attn_dist}
\end{figure*}




\section{Detailed Analysis of Attention in VLMs}
\label{sec:attention}

\subsection{Attention distribution}

\begin{figure*}[t]
  \centering

  \begin{subfigure}{\linewidth}
    \includegraphics[width=\linewidth]{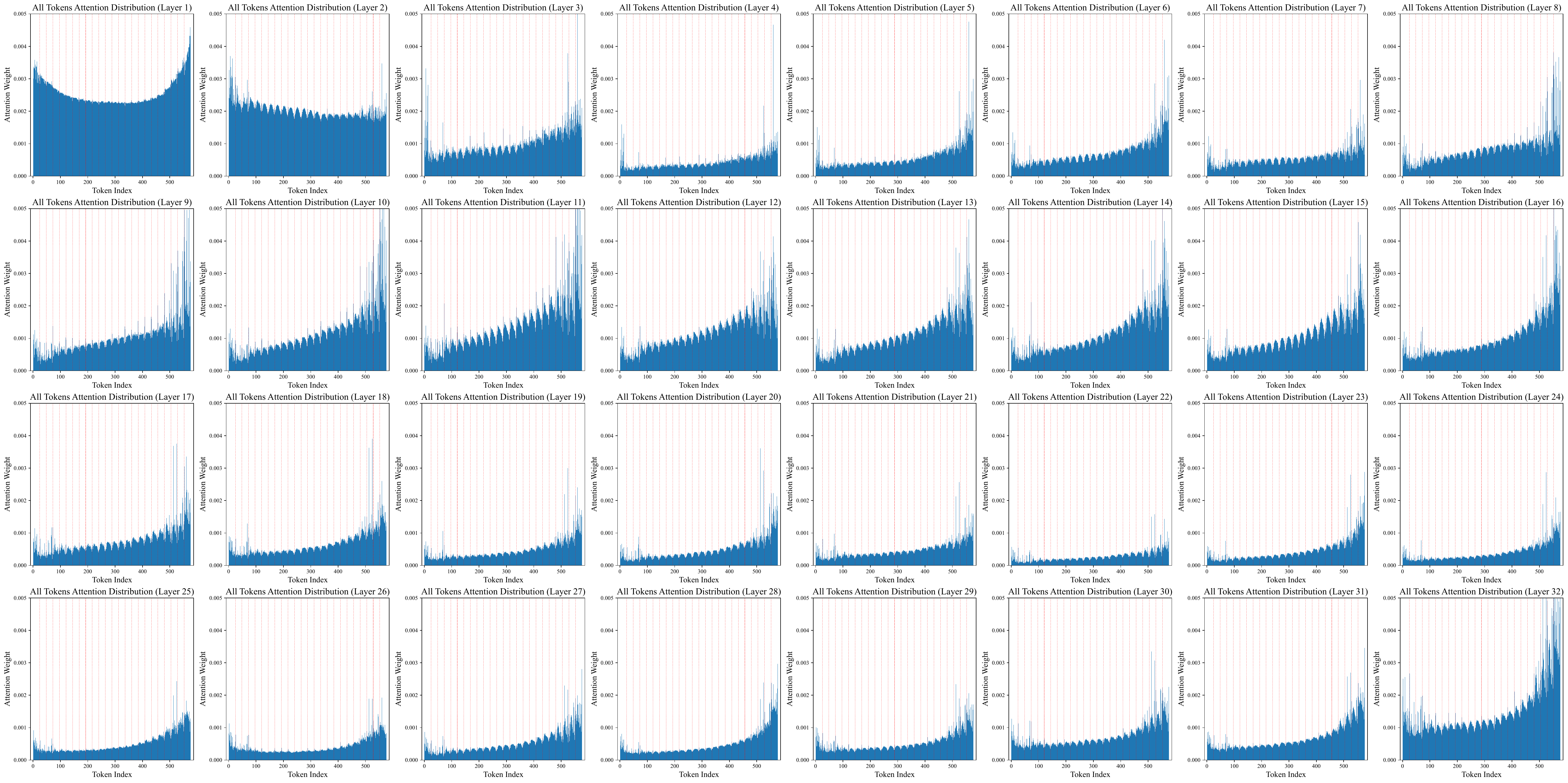}
    \label{fig:full_attn}
  \end{subfigure}
  
  \begin{subfigure}{\linewidth}
    \includegraphics[width=\linewidth]{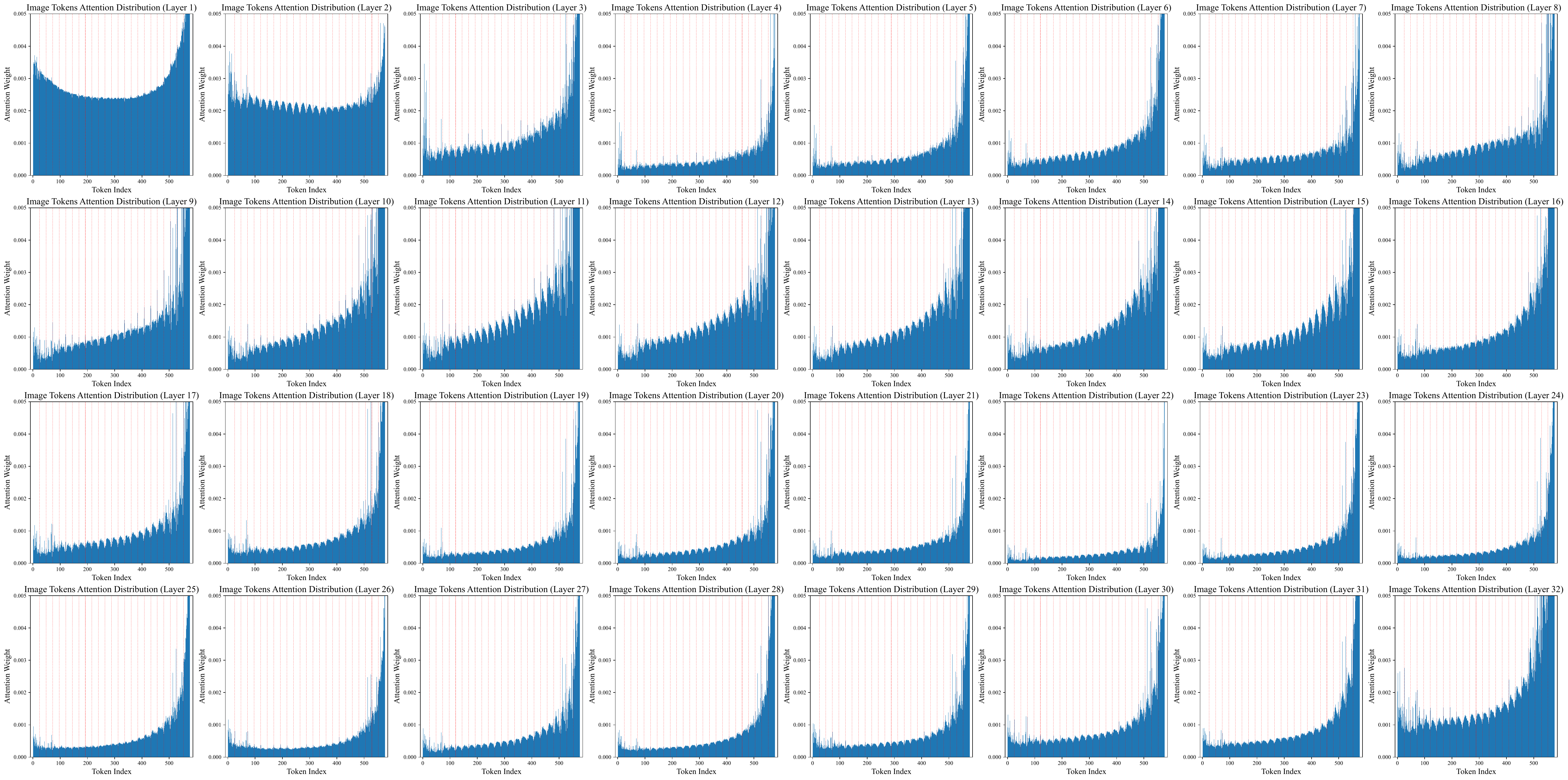}
    \label{fig:image_attn}
  \end{subfigure}
  
  \caption{Distribution of visual-text attention over visual token positions in LLaMA. The top rows display the average attention distribution across all tokens, while the bottom rows display the attention distribution from other visual tokens. Each type of attentions include results from all 32 layers of the 7B language model. The red vertical dashed lines indicate the length of each row in the input image (24 for CLIP-ViT-L-14-336px used in LLaVA-1.5).}
  \label{fig:full_image_attn_dist}
  \vspace{-5mm}
\end{figure*}

\begin{figure*}[t]
  \centering

  \begin{subfigure}{\linewidth}
    \includegraphics[width=\linewidth]{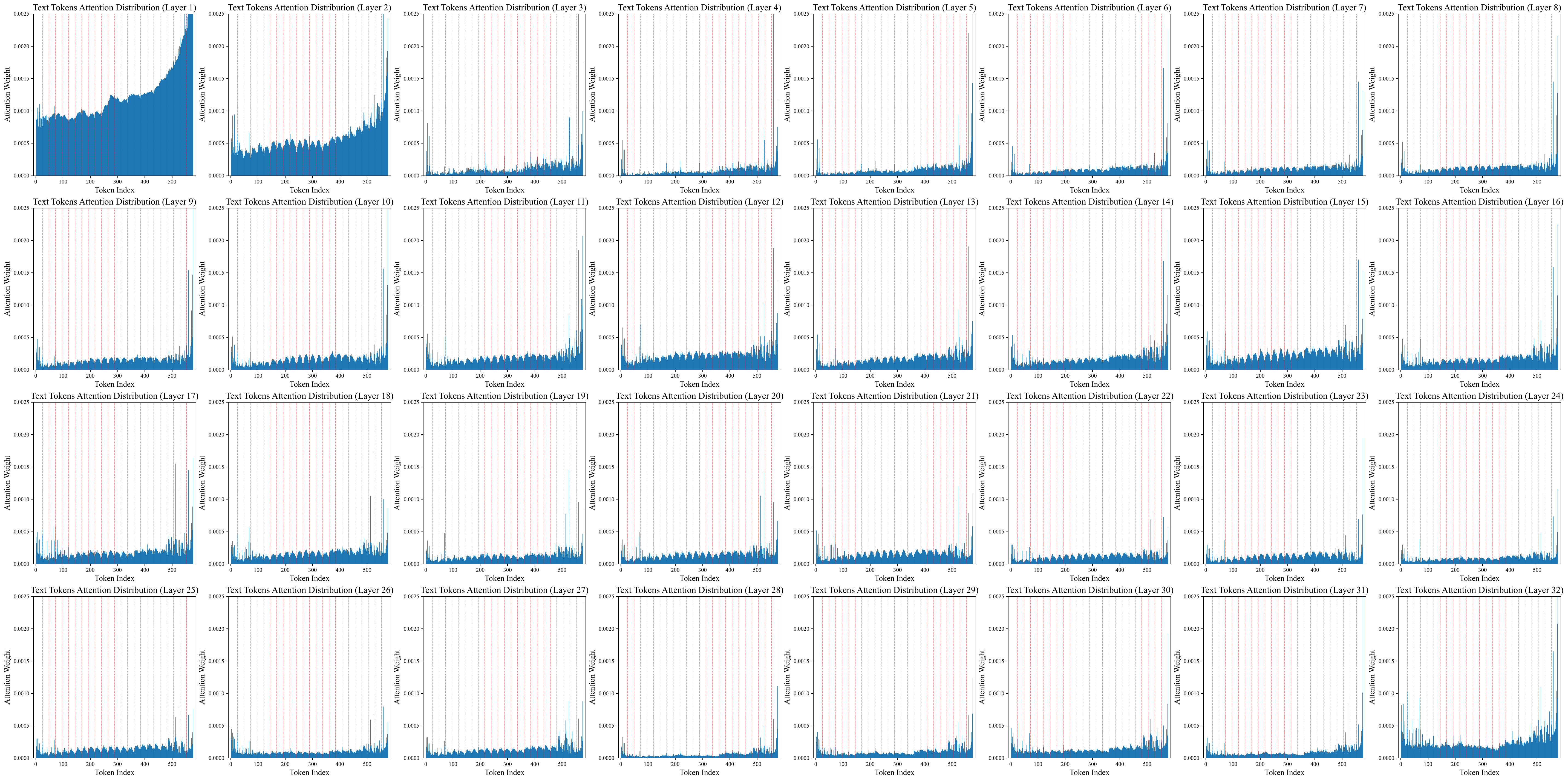}
    \label{fig:text_attn_map}
  \end{subfigure}
  
  \begin{subfigure}{\linewidth}
    \includegraphics[width=\linewidth]{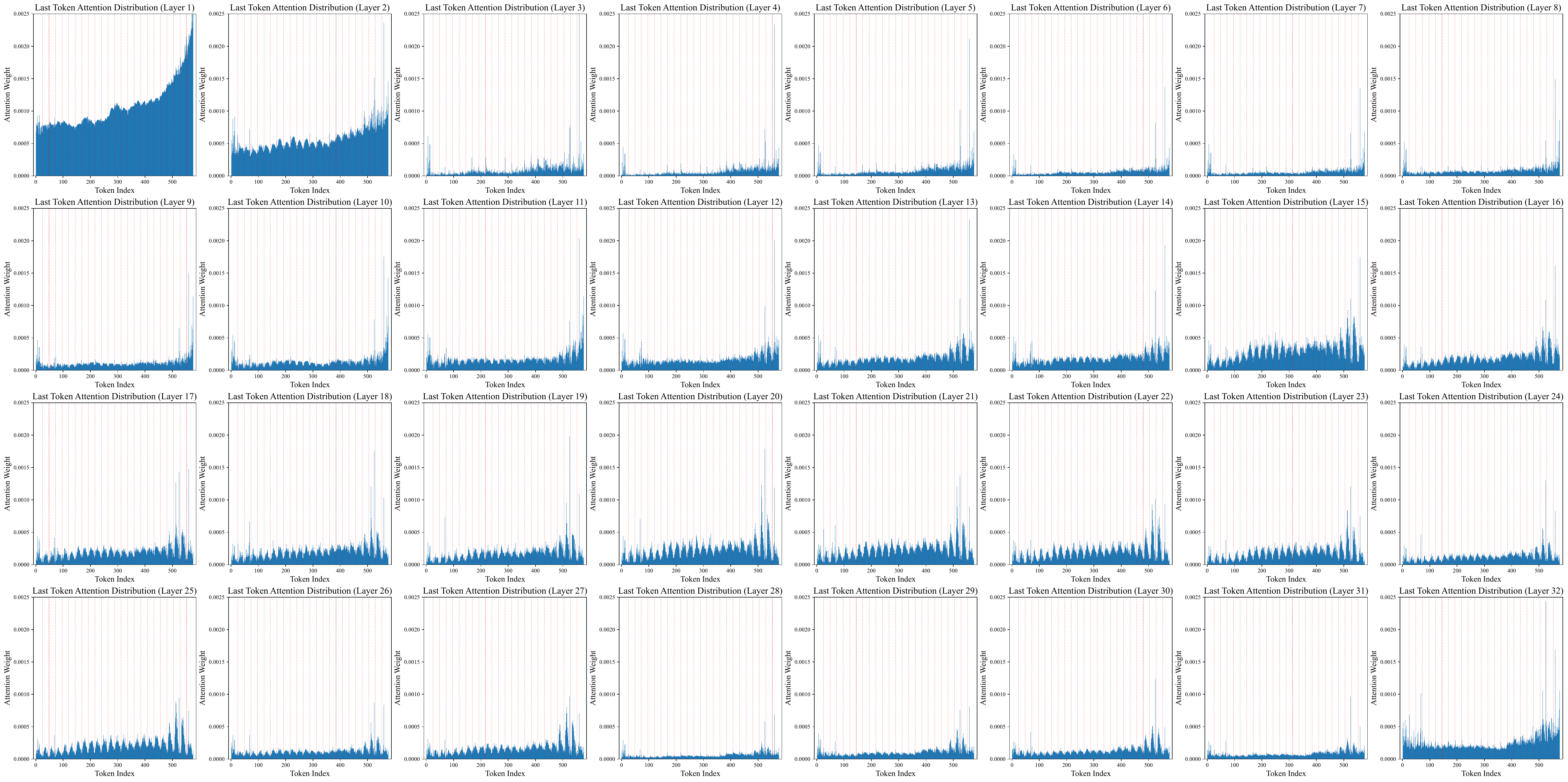}
    \label{fig:last_attn_map}
  \end{subfigure}

  \vspace{-3mm}
  \caption{Distribution of visual-text attention over visual token positions in LLaMA. The top rows display the attention distribution from language instruction tokens, while the bottom rows display the attention distribution of the last token, which is also used to predict the next token. Each type of attentions include results from all 32 layers of the 7B language model. The red vertical dashed lines indicate the length of each row in the input image (24 for CLIP-ViT-L-14-336px used in LLaVA-1.5).}
  \label{fig:text_last_attn_dist}
  \vspace{-5mm}
\end{figure*}

We first present the distribution of visual attention in CLIP. As shown in \cref{fig:clip_attn_dist}, the left two subplots show the visual attention in the penultimate layer of CLIP. The visual tokens used in LLaVA-1.5 also come from this layer, which retains more local features and image details. The right two subplots show the visual attention in the final layer of CLIP, which serves as the output layer. In the penultimate layer, attention from the [CLS] token is more concentrated compared to attention from other visual tokens, primarily focusing on regions closer to the image center. The [CLS] attention in the final layer is similar to that in the penultimate layer, but attention from other visual tokens becomes uniformly distributed due to the lack of supervision signals. Based on these observations, we adpot the [CLS] attention from the penultimate layer for visual token pruning.

\cref{fig:full_image_attn_dist} and \cref{fig:text_last_attn_dist} show the distribution of visual-text attention in the 32 layers of the 7B LLaMA language model, focusing on attention from all tokens, other visual tokens, language instruction tokens, and the last token. Unlike the attention distributions in the visual encoder, these visual-text attention distributions exhibit a clear trend of increasing intensity with larger token indices, which termed \textit{attention shift} in the main text. Pruning visual tokens based on such attention leads to significant performance degradation, especially at high reduction ratios. This shift phenomenon is consistently observed across all types of visual-text attention. Notably, attention from the text tokens is significantly weaker than from the visual part, especially after the second layer, corroborating the inefficient visual attention phenomenon identified in FastV~\cite{chen2024fastv} and providing sufficient motivation for visual token pruning. Additionally, we observe that the attention distribution in the first 2 layers differs noticeably from the subsequent 30 layers. Deeper analysis of this distinction and how to leverage it to improve VLM performance is left as future work.

\subsection{Attention intensity}

\begin{figure*}[t]
  \centering
  \includegraphics[width=\linewidth]{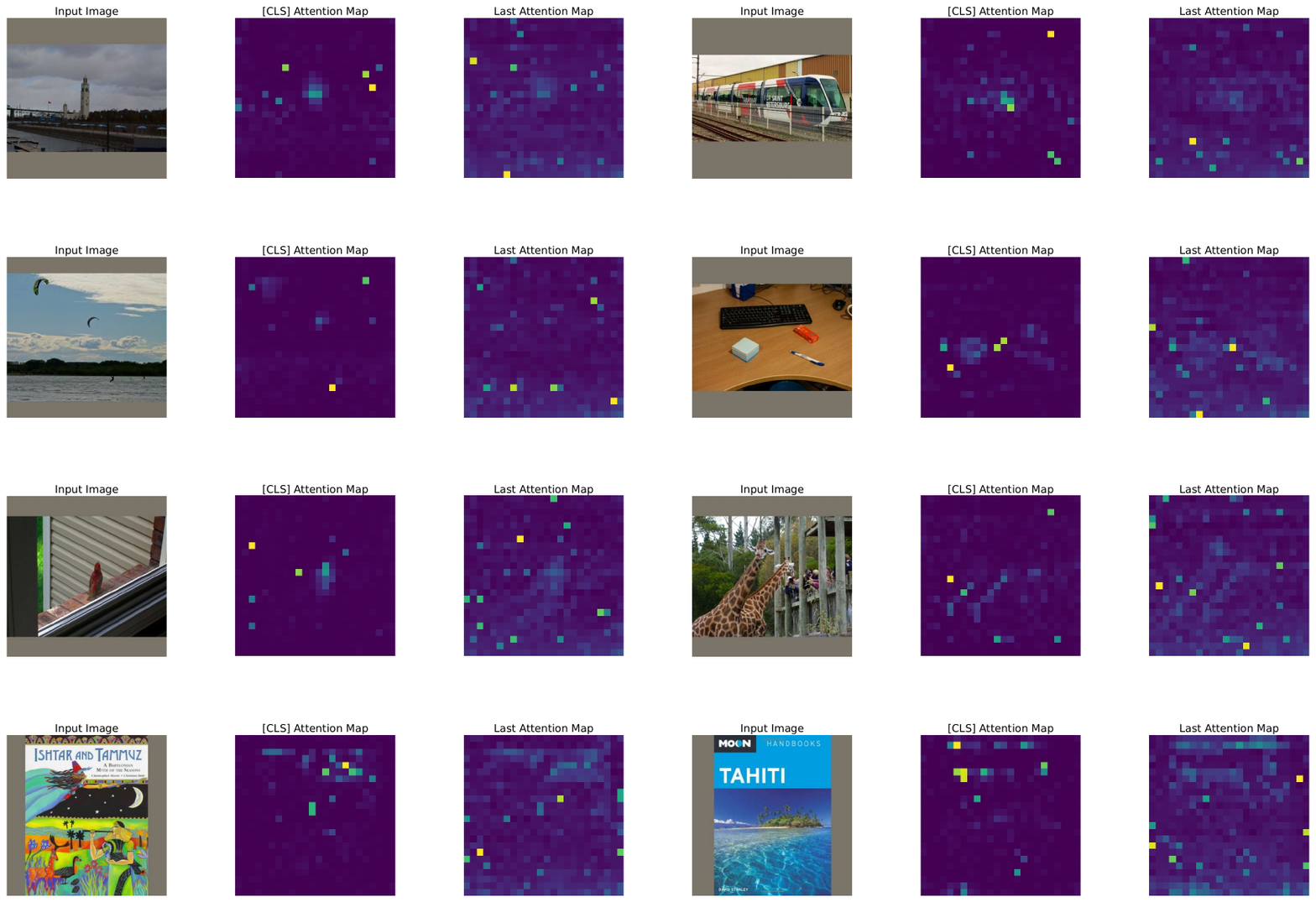}
  \caption{Visualizations of attention maps from the [CLS] token in the visual encoder and the last token in the language model.}
  \label{fig:attn_intensity}
  \vspace{20mm}
\end{figure*}

We visualize the attention maps from the [CLS] token in the visual encoder and the last token in the language model for the same input images in \cref{fig:attn_intensity}. The attention from the [CLS] token is more concentrated, focusing on key foreground objects like Big Ben, train carriages, paragliders, sticky notes, the bird, giraffes, and the titles on book covers, as well as certain artifacts that encode global information. In contrast, the attention from the last token in the language model is more dispersed, spread across the entire input image. This indicates that it includes more noise, making it less effective for accurately evaluating the importance of visual tokens. This discrepancy highlights a degree of misalignment between the visual and language modalities in existing vision-language models. Addressing this misalignment to improve VLM performance on multi-modal understanding tasks remains a direction for future work.

\section{Efficiency Analysis with FlashAttention}
\label{sec:efficiency}

In \cref{tab:flashattention-1.5-7b}, \cref{tab:flashattention-1.5-13b} and \cref{tab:flashattention-1.6-7b}, we compare the computational efficiency between FastV and our VisPruner under LLaVA-1.5-7B, LLaVA-1.5-13B, and LLaVA-NeXT-7B, respectively. Unlike FastV, which prune visual token within the LLM, VisPruner prunes tokens before the LLM, enabling compatibility with FlashAttention. This design results in significantly higher efficiency. Note that the original implementation of SDPA also includes FlashAttention, so its computational efficiency is comparable to that of FlashAttention2, with only slight differences. All efficiency analyses are performed on a single NVIDIA A100-80GB GPU, evaluated using the POPE benchmark.

\begin{table*}[t]
  \centering
  \resizebox{0.9\linewidth}{!}{
    \begin{tabular}{l|c|c|c|c|c|c|c}
    \toprule
    Method                      & Reduction             & \# Token             & FLOPs (T)                      & Storage (MB)                     & GPU Memory (GB)                 & CUDA Time (ms)  & \multicolumn{1}{l}{Accuracy (\%)} \\
    \midrule
    LLaVA-1.5-7B                & 0\%                   & 576                  & 8.02                           & 288.00                           & 14.68                           & 107.26          & 85.88                             \\
    FastV                       & \multirow{3}{*}{25\%} & \multirow{3}{*}{432} & 6.20                           & 220.50                           & 14.58                           & 107.09          & 85.29                             \\
    VisPruner (sdpa)            &                       &                      & \multirow{2}{*}{\textbf{6.08}} & \multirow{2}{*}{\textbf{216.00}} & \multirow{2}{*}{\textbf{14.51}} & 101.95          & \textbf{85.92}                    \\
    VisPruner (flash attention) &                       &                      &                                &                                  &                                 & \textbf{101.35} & 85.87                             \\
    \midrule
    FastV                       & \multirow{3}{*}{50\%} & \multirow{3}{*}{288} & 4.40                           & 153.00                           & 14.52                           & 99.72           & 82.45                             \\
    VisPruner (sdpa)            &                       &                      & \multirow{2}{*}{\textbf{4.16}} & \multirow{2}{*}{\textbf{144.00}} & \multirow{2}{*}{\textbf{14.46}} & 93.57           & \textbf{86.20}                    \\
    VisPruner (flash attention) &                       &                      &                                &                                  &                                 & \textbf{92.26}  & 86.16                             \\
    \midrule
    FastV                       & \multirow{3}{*}{75\%} & \multirow{3}{*}{144} & 2.62                           & 85.50                            & 14.52                           & 94.67           & 73.74                             \\
    VisPruner (sdpa)            &                       &                      & \multirow{2}{*}{\textbf{2.26}} & \multirow{2}{*}{\textbf{72.00}}  & \multirow{2}{*}{\textbf{14.44}} & 85.06           & \textbf{83.46}                    \\
    VisPruner (flash attention) &                       &                      &                                &                                  &                                 & \textbf{84.03}  & 83.42                             \\
    \midrule
    FastV                       & \multirow{3}{*}{90\%} & \multirow{3}{*}{58}  & 1.57                           & 45.31                            & 14.64                           & 90.48           & 57.30                             \\
    VisPruner (sdpa)            &                       &                      & \multirow{2}{*}{\textbf{1.13}} & \multirow{2}{*}{\textbf{29.00}}  & \multirow{2}{*}{\textbf{14.54}} & 79.11           & \textbf{75.85}                    \\
    VisPruner (flash attention) &                       &                      &                                &                                  &                                 & \textbf{77.44}  & 75.82                             \\
    \midrule
    FastV                       & \multirow{3}{*}{95\%} & \multirow{3}{*}{29}  & 1.22                           & 31.72                            & 14.63                           & 89.31           & 35.47                             \\
    VisPruner (sdpa)            &                       &                      & \multirow{2}{*}{\textbf{0.76}} & \multirow{2}{*}{\textbf{14.50}}  & \multirow{2}{*}{\textbf{14.54}} & 78.09           & \textbf{67.24}                    \\
    VisPruner (flash attention) &                       &                      &                                &                                  &                                 & \textbf{77.15}  & 67.22                             \\
    \bottomrule
    \end{tabular}
  }
  \caption{Efficiency comparison between FastV and VisPruner under LLaVA-1.5-7B.}
  \label{tab:flashattention-1.5-7b}
\end{table*}

\begin{table*}[t]
  \centering
  \resizebox{0.9\linewidth}{!}{
    \begin{tabular}{l|c|c|c|c|c|c|c}
    \toprule
    Method                      & Reduction             & \# Token             & FLOPs (T)                       & Storage (MB)                     & GPU Memory (GB)                 & CUDA Time (ms)  & \multicolumn{1}{l}{Accuracy (\%)} \\
    \midrule
    LLaVA-1.5-13B               & 15.28                 & 450.00               & 26.98                           & 156.64                           & 85.99                           & 107.26          & 85.88                             \\
    FastV                       & \multirow{3}{*}{25\%} & \multirow{3}{*}{432} & 11.69                           & 343.13                           & 26.61                           & 151.66          & 85.86                             \\
    VisPruner (sdpa)            &                       &                      & \multirow{2}{*}{\textbf{11.50}} & \multirow{2}{*}{\textbf{337.50}} & \multirow{2}{*}{\textbf{26.58}} & 138.36          & \textbf{86.73}                    \\
    VisPruner (flash attention) &                       &                      &                                 &                                  &                                 & \textbf{137.95} & 86.72                             \\
    \midrule
    FastV                       & \multirow{3}{*}{50\%} & \multirow{3}{*}{288} & 8.14                            & 236.25                           & 26.40                           & 137.83          & 85.15                             \\
    VisPruner (sdpa)            &                       &                      & \multirow{2}{*}{\textbf{7.76}}  & \multirow{2}{*}{\textbf{225.00}} & \multirow{2}{*}{\textbf{26.31}} & 124.13          & \textbf{86.05}                    \\
    VisPruner (flash attention) &                       &                      &                                 &                                  &                                 & \textbf{123.33} & 86.04                             \\
    \midrule
    FastV                       & \multirow{3}{*}{75\%} & \multirow{3}{*}{144} & 4.62                            & 129.38                           & 26.40                           & 123.69          & 79.43                             \\
    VisPruner (sdpa)            &                       &                      & \multirow{2}{*}{\textbf{4.05}}  & \multirow{2}{*}{\textbf{112.50}} & \multirow{2}{*}{\textbf{26.29}} & 104.81          & \textbf{83.10}                    \\
    VisPruner (flash attention) &                       &                      &                                 &                                  &                                 & \textbf{103.57} & 83.09                             \\
    \midrule
    FastV                       & \multirow{3}{*}{90\%} & \multirow{3}{*}{58}  & 2.53                            & 65.70                            & 26.40                           & 114.98          & 67.26                             \\
    VisPruner (sdpa)            &                       &                      & \multirow{2}{*}{\textbf{1.86}}  & \multirow{2}{*}{\textbf{45.31}}  & \multirow{2}{*}{\textbf{26.27}} & 94.80           & \textbf{74.71}                    \\
    VisPruner (flash attention) &                       &                      &                                 &                                  &                                 & \textbf{94.04}  & 74.66                             \\
    \midrule
    FastV                       & \multirow{3}{*}{95\%} & \multirow{3}{*}{29}  & 1.83                            & 44.19                            & 26.55                           & 114.25          & 49.83                             \\
    VisPruner (sdpa)            &                       &                      & \multirow{2}{*}{\textbf{1.12}}  & \multirow{2}{*}{\textbf{22.66}}  & \multirow{2}{*}{\textbf{26.43}} & 94.48           & \textbf{65.90}                    \\
    VisPruner (flash attention) &                       &                      &                                 &                                  &                                 & \textbf{93.73}  & 65.79                             \\
    \bottomrule
    \end{tabular}
  }
  \caption{Efficiency comparison between FastV and VisPruner under LLaVA-1.5-13B.}
  \label{tab:flashattention-1.5-13b}
\end{table*}

\begin{table*}[t]
  \centering
  \resizebox{0.9\linewidth}{!}{
    \begin{tabular}{l|c|c|c|c|c|c|c}
    \toprule
    Method                      & Reduction             & \# Token              & FLOPs (T)                       & Storage (MB)                      & GPU Memory (GB)                 & CUDA Time (ms)  & \multicolumn{1}{l}{Accuracy (\%)} \\
    \midrule
    LLaVA-NeXT-7B               & 0\%                   & 2880                  & 43.58                           & 1440.00                           & 17.04                           & 313.04          & 86.77                             \\
    \midrule
    FastV                       & \multirow{3}{*}{25\%} & \multirow{3}{*}{2160} & 33.05                           & 1102.50                           & 16.95                           & 262.49          & 86.63                             \\
    VisPruner (sdpa)            &                       &                       & \multirow{2}{*}{\textbf{32.35}} & \multirow{2}{*}{\textbf{1080.00}} & \multirow{2}{*}{\textbf{16.33}} & 246.38          & \textbf{86.69}                    \\
    VisPruner (flash attention) &                       &                       &                                 &                                   &                                 & \textbf{232.07} & 86.66                             \\
    \midrule
    FastV                       & \multirow{3}{*}{50\%} & \multirow{3}{*}{1440} & 23.04                           & 765.00                            & 16.95                           & 201.68          & 85.73                             \\
    VisPruner (sdpa)            &                       &                       & \multirow{2}{*}{\textbf{21.66}} & \multirow{2}{*}{\textbf{720.00}}  & \multirow{2}{*}{\textbf{15.11}} & 176.80          & \textbf{87.06}                    \\
    VisPruner (flash attention) &                       &                       &                                 &                                   &                                 & \textbf{170.30} & 87.02                             \\
    \midrule
    FastV                       & \multirow{3}{*}{75\%} & \multirow{3}{*}{720}  & 13.53                           & 427.50                            & 16.95                           & 147.93          & 82.68                             \\
    VisPruner (sdpa)            &                       &                       & \multirow{2}{*}{\textbf{11.51}} & \multirow{2}{*}{\textbf{360.00}}  & \multirow{2}{*}{\textbf{14.80}} & 119.87          & \textbf{86.50}                    \\
    VisPruner (flash attention) &                       &                       &                                 &                                   &                                 & \textbf{116.79} & 86.46                             \\
    \midrule
    FastV                       & \multirow{3}{*}{90\%} & \multirow{3}{*}{290}  & 8.10                            & 226.56                            & 16.95                           & 117.33          & 70.77                             \\
    VisPruner (sdpa)            &                       &                       & \multirow{2}{*}{\textbf{5.71}}  & \multirow{2}{*}{\textbf{145.00}}  & \multirow{2}{*}{\textbf{14.70}} & 87.36           & \textbf{81.17}                    \\
    VisPruner (flash attention) &                       &                       &                                 &                                   &                                 & \textbf{85.26}  & 81.12                             \\
    \midrule
    FastV                       & \multirow{3}{*}{95\%} & \multirow{3}{*}{145}  & 6.30                            & 158.59                            & 16.95                           & 112.19          & 49.36                             \\
    VisPruner (sdpa)            &                       &                       & \multirow{2}{*}{\textbf{3.80}}  & \multirow{2}{*}{\textbf{72.50}}   & \multirow{2}{*}{\textbf{14.70}} & 78.18           & \textbf{74.77}                    \\
    VisPruner (flash attention) &                       &                       &                                 &                                   &                                 & \textbf{77.66}  & 74.72                             \\
    \bottomrule
    \end{tabular}
  }
  \caption{Efficiency comparison between FastV and VisPruner under LLaVA-NeXT-7B.}
  \label{tab:flashattention-1.6-7b}
\end{table*}



\end{document}